\documentclass[lettersize,journal]{IEEEtran}
\usepackage{amsmath,amsfonts}
\usepackage{algorithmic}
\usepackage{algorithm}
\usepackage{array}
\usepackage{stmaryrd}
\usepackage{url}
\usepackage{multirow}
\usepackage{arydshln}
\usepackage{booktabs}
\usepackage[english]{babel}
\usepackage{xcolor}
\usepackage{amsthm}
\usepackage{bm}
\usepackage{verbatim}
\usepackage{graphicx}

\usepackage{orcidlink}
\usepackage{subcaption}
\usepackage[numbers]{natbib}

\usepackage[export]{adjustbox}
\newtheorem{theorem}{Theorem}[section]

\DeclareMathOperator*{\argmin}{argmin}

\begin{document}

\title{Privacy-Preserving Parameter-Efficient Fine-Tuning for Large Language Model Services}

\author{Yansong Li\orcidlink{0009-0004-4694-8419
}, Zhixing Tan\orcidlink{0000-0002-2426-6220}, Paula Branco\orcidlink{0000-0002-9917-3694} and Yang Liu\orcidlink{0000-0002-3087-242X},~\IEEEmembership{Senior Member,~IEEE}
\thanks{This work was supported by the National Natural Science Foundation of China under Grant 62006138 and Grant 62236011, and in part by the National Social Science Fund of China under Grant 20\&ZD279. Supported by Zhongguancun Laboratory. (\emph{Corresponding authors: ZhixingTan; Yang Liu}.)}
\thanks{Yansong Li is with the University of Ottawa, Ottawa, Canada (email: yli627@uottawa.ca)}
\thanks{Zhixing Tan is with Zhongguancun Laboratory, Beijing, P.R.China (email: zxtan@zgclab.edu.cn).}
\thanks{Paula Branco is with the School of Electrical Engineering and Computer Science, University of Ottawa, Ottawa, Canada (email: pbranco@uottawa.ca).}
\thanks{Yang Liu is with the Department of Computer Science and Technology, Tsinghua University, Beijing, China; Institute for AI Industry Research (AIR), Tsinghua University, Beijing, China (e-mail: liuyang2011@tsinghua.edu.cn).}
}

\markboth{IEEE Transactions on Audio, Speech, and Language Processing}
{Li \MakeLowercase{\textit{et al.}}: Privacy-Preserving Parameter-Efficient Fine-Tuning for Large Language Model Services}

\maketitle

\begin{abstract}
Parameter-Efficient Fine-Tuning (PEFT) provides a practical way for users to customize Large Language Models (LLMs) with their private data in LLM service scenarios. However, the inherently sensitive nature of private data demands robust privacy preservation measures during the customization of LLM services to ensure data security, maintain user trust, and comply with stringent regulatory standards. Based on PEFT, we propose P\underline{r}iv\underline{a}cy-Preserving \underline{P}arameter-Efficient Fine-\underline{T}uning (\textsc{rapt}), a framework that offers privacy protection for LLM services. \textsc{rapt} adopts a local privacy approach, enabling users to privatize their data locally using a text-to-text local differential privacy mechanism. Since PEFT performs poorly when directly trained on privatized data, we introduce a novel privatized token reconstruction task that is trained jointly with the downstream task, allowing LLMs to learn better task-dependent representations. Despite the simplicity of our framework, experiments show that \textsc{rapt} achieves competitive performance across tasks while providing privacy guarantees against adversaries.
\end{abstract}


\section{Introduction}
\IEEEPARstart{R}{ecent} years have witnessed the tremendous success of Large Language Models (LLMs)~\cite{devlin2018bert,brown2020language}. As the size of LLMs continues to grow, it becomes increasingly difficult for individual users to deploy and run them locally. Consequently, recent LLMs are often released as cloud services. To enable users and developers to customize these LLM services, service providers typically offer fine-tuning APIs, which are generally based on Parameter-Efficient Fine-Tuning (PEFT) methods~\cite{hu2021lora,li2021prefix,lester2021power}. PEFT provides a straightforward and practical solution for adapting LLMs to downstream tasks while achieving competitive performance compared to full fine-tuning~\cite{lester2021power}.

\begin{figure*}[!t]
\centering 
  \begin{subfigure}[b]{0.45\textwidth}
  \centering
  \includegraphics[width=\textwidth]{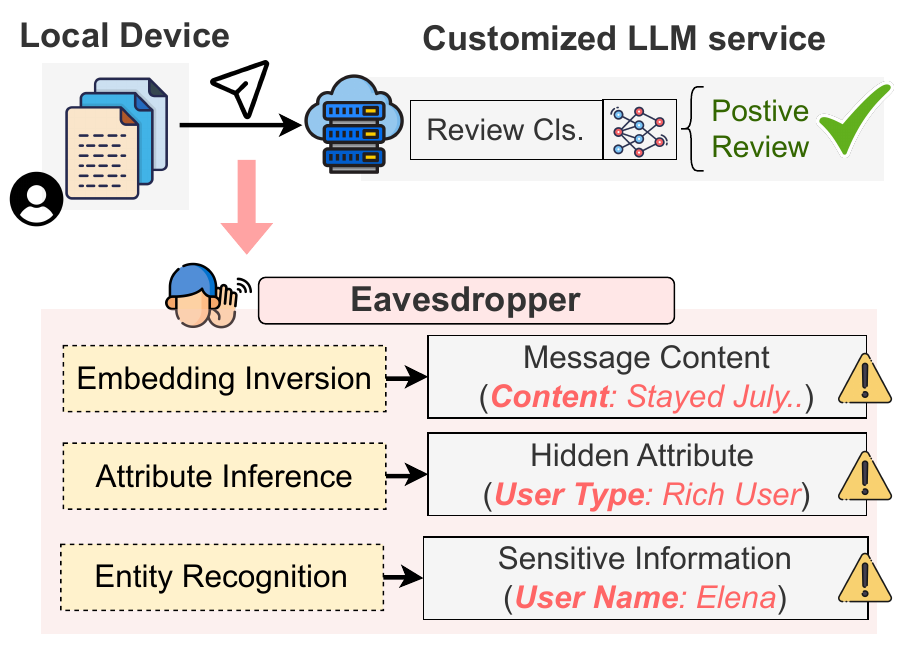}
  \caption{PEFT without Privacy Protection.}
  \end{subfigure}
  \hfill
  \begin{subfigure}[b]{0.535\textwidth}
  \centering
  \includegraphics[width=\textwidth]{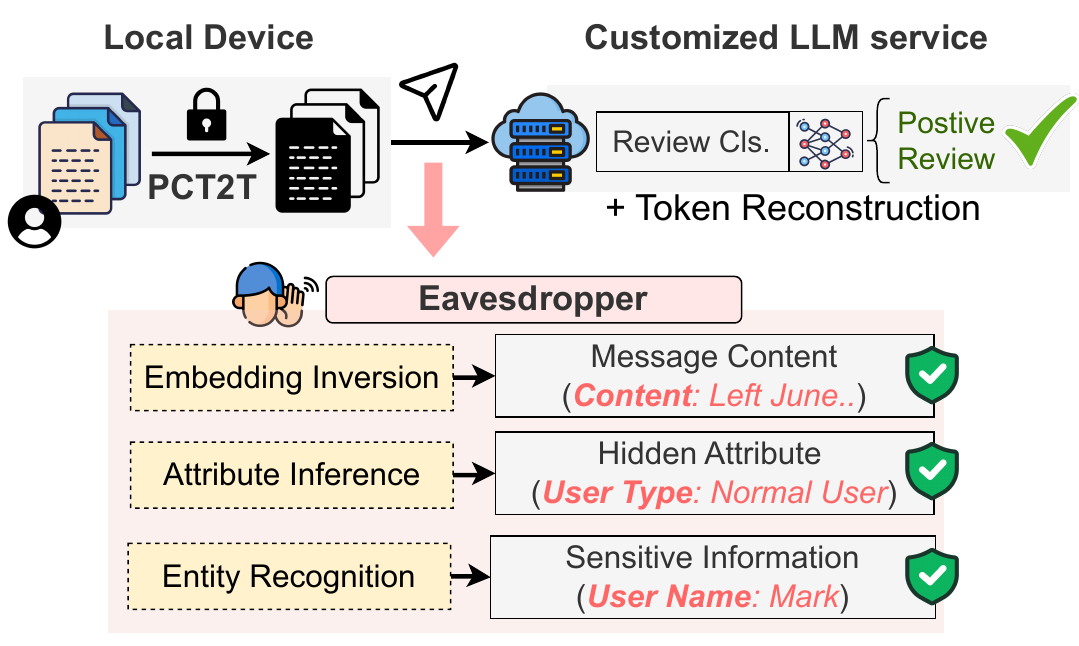}
  \caption{Our \textsc{rapt} framework.}
  \end{subfigure}
  \caption{Comparison between PEFT without privacy protection and our proposed \textsc{rapt} framework. In \textsc{rapt}, users can privatize their data locally using PCT2T. The customized LLM from the service provider is trained with privatized token reconstruction and attempts to recover the original input representations from their privatized counterparts during inference. This framework offers strong privacy protection against potential eavesdropping attacks, such as attribute inference attacks.}\label{fig:overview}
\end{figure*}

Although PEFT APIs provide a practical way for users to customize and utilize LLM services, the sensitive nature of private data raises concerns about potential leakage, particularly with respect to Personally Identifiable Information (PII) and Protected Health Information (PHI). Both categories of data are fundamental to the protection of individual privacy rights~\cite{martin2016measuring} and are subject to stringent legal regulations~\cite{gdpr2016general,pardau2018california}. To customize and use LLM services through PEFT APIs, users must first upload their data to the service provider. However, it is well known that input texts or even embedding representations can leak private information to various adversaries~\cite{coavoux2018privacy}. Therefore, the effective preservation of privacy is essential to ensure the trustworthiness of LLM services.

However, ensuring privacy preservation in the context of LLM services is challenging. First, providing privacy protection for sensitive data in LLM services is not straightforward. Existing works on LLM privacy preservation~\cite{yu2021differentially,shi2021selective,anil2021large,hoory2021learning,li2021large,shi2022just} primarily focus on \emph{centralized privacy settings}, which rely on the service provider to safeguard users from privacy leakage. Such settings may be inadequate in scenarios involving an honest-but-curious service provider or a middle eavesdropper~\cite{lyu2020differentially}. Second, imposing privacy protection often degrades the performance of downstream tasks, resulting in the well-known \emph{privacy-utility trade-off}~\cite{sreekumar2019optimal, ye2024initialization, zhang2024utility}. Previous studies have found that text privacy mechanisms struggle to preserve semantics while safeguarding private information~\cite{arnold-etal-2023-guiding}, and PEFT methods lack robustness against word perturbations~\cite{yang2022robust}. Consequently, customizing LLMs with PEFT is likely to incur significant performance degradation when privacy protection is applied.

To tackle the challenges mentioned above, we propose P\underline{r}iv\underline{a}cy-Preserving \underline{P}arameter-Efficient Fine-\underline{T}uning (\textsc{rapt}), a framework for customizing and utilizing LLM services with privacy preservation. For this purpose, \textsc{rapt} incorporates local privacy settings that allow users to privatize their data directly on their devices. Specifically, \textsc{rapt} utilizes Text-to-Text (T2T) privatization~\cite{feyisetan2020privacy}, a form of local differential privacy (LDP)~\cite{evfimievski2003limiting}, which first injects random perturbations into word representations and then substitutes words based on a nearest neighbor search in the representation space. As sensitive data is already privatized before being sent to the service provider, users can protect their data against various eavesdropping attacks, such as embedding inversion, attribute inference, and Named Entity Recognition (NER) attacks. To mitigate performance degradation when applying T2T privatization, \textsc{rapt} introduces an additional Part-of-Speech (POS) constraint, which restricts word substitutions within their original POS roles, thus achieving better grammatical coherence than the standard T2T mechanism. Moreover, we propose a novel privatized token reconstruction task, building on the findings that the masked language modeling objective can learn separable deep representations~\cite{voita2019bottom,mamou2020emergence,wettig2022should}, to further enhance the performance of \textsc{rapt} on downstream tasks. The objective of this task is to recover the original content of a privatized special token sequence from LLM representations. Unlike standard PEFT, \textsc{rapt} jointly tunes parameters for both the downstream task and the privatized token reconstruction task. In this way, \textsc{rapt} combines the lightweight and modular benefits of PEFT with improved utility (\emph{i.e.}, performance) through the use of POS constrained T2T (PCT2T) and privatized token reconstruction. Figure~\ref{fig:overview} provides a comparison between PEFT without privacy protection and our proposed \textsc{rapt} framework.

We conduct simulated privacy attacks and experiments on various tasks with different LLMs to evaluate the effectiveness of our proposed framework. The results demonstrate that \textsc{rapt} achieves competitive performance while providing privacy protection across different tasks and LLMs.

\section{Related Work}
Recent studies on LLM privacy protection can be broadly categorized into centralized and local approaches.

\paragraph{Centralized Approaches} Most existing works focus on a centralized privacy setting, relying on a central data curator to safeguard data from privacy leakage. Numerous studies have explored how to train privacy-preserving LLMs~\cite{carlini2021extracting,hoory2021learning,li2021large,yu2021large}, which are beyond the scope of this work. For instance, \citet{kerrigan2020differentially} utilizes differentially private fine-tuning to protect private data used in fine-tuning public language models. \citet{yu2021differentially} investigate privacy protection during the fine-tuning stage using lightweight methods such as \textsc{adapter}s~\cite{hu2021lora} and \textsc{prefix-tuning}~\cite{li2021prefix}. Our study differs from these works by focusing on protecting personal data in LLM services without assuming a central data curator.

\paragraph{Local Approaches} Local approaches offer a higher degree of privacy protection but often at the expense of utility. \citet{lyu2020differentially} explore maintaining model utility under local privacy constraints by utilizing a differentially private neural representation method. However, their approach solely addresses privacy protection during the inference stage. Conversely, several studies have proposed pre-training~\cite{qu2021pcf,hoory2021learning,ding2024delving} and fine-tuning~\cite{huang2020texthide,qu2021pcf,lukas2023analyzing,shen2023split,ye2024openfedllm} based methods that provide privacy protection during both training and inference stages. These methods require fine-tuning the entire model on privatized data, which is costly when customizing LLM services for many users. Our approach also adopts a local privacy setting and supports privacy protection during both fine-tuning and inference but remains lightweight. Additionally, we introduce a novel privatized token reconstruction task to enhance the performance of prompt tuning methods when training and inferring with privatized data.

\section{Approach}
In this section, we provide a detailed description of \textsc{rapt}. We begin by introducing Text-to-Text (T2T) privatization, which allows users to privatize their data locally. Next, we describe how to customize and utilize LLM services with privatized data using our method.

\subsection{Text-to-Text Privatization}~\label{sec:privacy}
Users of LLM services may encounter data privacy risks due to eavesdropping attacks or honest-but-curious service providers. Therefore, adopting a local privacy setting is advantageous, \textbf{allowing users to perform data privatization locally on their devices}. In our framework, we utilize T2T privatization~\cite{feyisetan2020privacy}, as most LLMs operate with a text-to-text interface. T2T privatization is based on $d_\mathcal{X}$-privacy~\cite{chatzikokolakis2013broadening}, a distance-based relaxation of local differential privacy that is widely used to protect textual content~\cite{feyisetan2020privacy,qu2021pcf}.

Formally, for a given input set $\mathcal{X}$ and an output set $\mathcal{Y}$, $d(\cdot,\cdot)$ denote a distance function on $\mathcal{X}$. A randomized mechanism $M: \mathcal{X} \rightarrow \mathcal{Y}$ satisfies $d_{\mathcal{X}}$-privacy if and only if, for any two data points $\bm{x}, \bm{x}' \in \mathcal{X}$, the distributions of outputs from $M(\bm{x})$ and $M(\bm{x}')$ are bounded by:
\begin{align}
\frac{P[M(\bm{x})=\bm{y}]}{P[M(\bm{x}')=\bm{y}]} \le e^{\eta d(\bm{x},\bm{x}')},  \quad \forall y \in \mathcal{Y}.
\end{align}
where $\eta\ge 0$ is the \emph{privacy parameter}, which controls the degree of privacy protection.\footnote{To avoid confusion, we follow \citet{qu2021pcf} and use $\eta$ to distinguish $\epsilon$, which is a commonly used parameter in differential privacy literature.}

\subsubsection{The Original Text-to-Text Privatization}
The original T2T privatization approach introduced by \citet{feyisetan2020privacy} demonstrates that a privatized sequence $\hat{\mathbf{x}}$ can be obtained by systematically replacing each word $w_t$ in the sequence $\mathbf{x} = [w_1, \ldots, w_n]$ with a different word $\hat{w}_t$, thereby ensuring plausible deniability~\cite{bindschaedler2017plausible}. Specifically, using the $L_2$ distance and an embedding model $\bm{E} \in \mathbb{R}^{|V| \times d}$, where $|V|$ is the vocabulary size and $d$ is the embedding dimension, the process begins by applying $d_\mathcal{X}$-privacy to the word embedding $\bm{w}_t = \bm{E}(w_t)$, with $\bm{w}_t \in \mathbb{R}^d$. This is typically achieved by adding random noise $\bm{z}$ to $\bm{w}_t$, drawn from a distribution with density $p(\bm{z}) \propto \exp(-\eta \|\bm{z}\|)$. The noise is expressed as $\bm{z} = l\bm{v}$, where $l \sim \Gamma(d, \frac{1}{\eta})$ and $\bm{v}$ is sampled uniformly from the unit ball $\mathbb{B}^d$, yielding the privatized embedding $\hat{\bm{w}}_t = \bm{w}_t + \bm{z}$. The final step applies nearest neighbor search to map the noisy representation $\hat{\bm{w}}_t$ to a corresponding word $\hat{w}_t$, formalized as $
\hat{w}_t = \argmin_{w_k} \|\bm{E}(w_k) - \hat{\bm{w}}_t\|$.

\subsubsection{Part-of-Speech Constrained Text-to-Text Privatization}

A major weakness of the original text-to-text privatization is that the transformation treats each word in a sentence equally and overlooks both the syntactic role and semantics of the word, which may result in significant syntactic and semantic changes to the sentence. For example, the sentence ``I would like to eat a burger'' is transformed into ``It rogate calhoun to drive 25 pulitzer'' after T2T privatization with a moderate privacy budget, exhibiting both syntactic changes (\textit{e.g.}, ``a $\rightarrow$  25'') and semantic changes (\textit{e.g}., ``eat $\rightarrow$ drive'').

To alleviate the above issue, we propose a Part-of-Speech constrained text-to-text privatization (PCT2T). Unlike the original T2T, which adds random noise to all words in a sentence, PCT2T only adds random noise to words belonging to a predefined set of POS categories and transforms the perturbation of representations into words within the same POS category. The benefits of PCT2T are twofold. First, with the additional POS constraint, the syntactic structure of the sentence remains the same as the original sentence after transformation. Second, as only words in selected POS categories are transformed, the transformed sentence is more semantically similar to the original sentence. By carefully choosing POS categories, we can significantly improve the utility while maintaining nearly the same degree of privacy protection.

Technically speaking, we first mark the word boundary before applying tokenization. After POS tagging, for each word $w_t$ in a selected POS category $\mathcal{C}$, the representation $\bm{w}_t$ of the word $w_t$ is calculated as $\bm{w}_t = \textrm{Mean}(\{\bm{w}_k\}_{w_k \in \textrm{Tok}(w_t)})$, where $\textrm{Tok}(w_t)$ denotes the set of tokens of $w_t$ after tokenization. Finally, the projected word $\hat{w}_t$ is identified by using nearest neighbor search:
\begin{align}
\hat{w}_t = \argmin_{w_k} \| \bm{E}_{C}(w_k) - \bm{w}_t \|, \quad \text{subject to } w_k \in  \mathcal{C},
\end{align}
where $\bm{E}_C$ denotes the embedding for the POS category $\mathcal{C}$ calculated from $\bm{E}$. The procedure for calculating a word embedding in $\bm{E}_C$ is the same as that for calculating $\bm{w}_t$ mentioned above.

\subsection{LLM Customization}
Being lightweight and modular, PEFT methods are suitable choices for customizing LLM services. We assume that \textbf{users customize LLM services through APIs by uploading their private training data to service providers}. For simplicity, we assume that the service provider uses \textsc{prompt tuning}~\cite{lester2021power} for LLM customization, although other PEFT methods are equally applicable.

\subsubsection{Prompt Tuning}
We follow the notation of \citet{lester2021power}, although variants such as \textsc{prefix-tuning}~\cite{li2021prefix} and \textsc{p-tuning}~\cite{liu2021p} also exist. \textsc{prompt tuning} uses continuous virtual tokens to adapt LLMs, and these tokens are learned through gradient descent~\cite{li2021prefix,lester2021power}. 

Let $f$ denote the backbone LLM model for the LLM service, and let $h$ be the hidden size of the LLM model. A prompt $\bm{T} \in \mathbb{R}^{l \times h}$ of length $l$ is a sequence of continuous vectors $[\bm{t}_1, \ldots, \bm{t}_l]$, where $\bm{t}_i \in \mathbb{R}^h$ is a tunable virtual token embedding. Given an input sequence $\mathbf{x} = [w_1, \ldots, w_n]$, the prompt is prepended to $\bm{X} = [\bm{w}_1, \ldots, \bm{w}_n]$, which represents the embeddings of an input sequence $\mathbf{x}$. We then obtain a sequence of activations using the LLM:
\begin{align}
\bm{H} = f(\llbracket \bm{T}; \bm{X} \rrbracket),
\end{align}
where $\bm{H} \in \mathbb{R}^{(l+n) \times h}$ is the sequence of activations, and $\llbracket \dots \rrbracket$ denotes the concatenation operator. The activations $\bm{H}$ are used to predict task-specific labels, for example via a language modeling head~\cite{liu2021p}.

\subsubsection{Privatized Token Reconstruction}

Unfortunately, directly tuning parameters on privatized data with PEFT significantly degrades the performance of the LLM on downstream tasks, even under weak privacy protections. This suggests that PEFT methods are highly sensitive to the random perturbations introduced by T2T privatization.

Recent studies on language model representations~\cite{voita2019bottom,mamou2020emergence} found that the masked language modeling objective~\cite{devlin2018bert} is effective in learning separable deep representations. Inspired by these findings, we propose a novel privatized token reconstruction task, which is trained jointly with the downstream task. We expect that training with privatized token reconstruction will help LLMs learn better task-dependent representations, thereby improving their performance on downstream tasks while preserving privacy.
 
Privatized token reconstruction is similar to masked language modeling. However, directly reconstructing privatized inputs is infeasible, as it would require revealing the original input. To address this issue, we prepend a fixed sequence of tokens, referred to as ``\emph{plain tokens},'' to each input in the training data. Plain tokens consist of arbitrarily chosen tokens and can be safely sent to the service provider without concerns about privacy leakage.

\begin{figure}[t]
    \centering
    \includegraphics[width=0.45\textwidth]{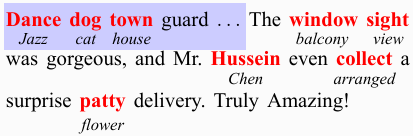}
    \caption{Example of privatized text. Changes introduced by privatization are highlighted in red, with the original words shown below their corresponding privatized counterparts. Plain tokens (abbreviated and shaded in blue) are randomly generated and are not required to be semantically meaningful. For readability, words are displayed in place of raw tokens.}
    \label{fig:train_sample}
\end{figure}

Formally, let $\mathbf{k} = [k_1, \ldots, k_m]$ denotes the plain tokens and $\mathcal{D} = \{ \langle \mathbf{x}_i, y_i \rangle \mid i = 1, \ldots, |\mathcal{D}| \}$ be the set of training examples, where $|\mathcal{D}|$ denotes the size of the training set and $y_i$ denotes the ground-truth label, the user first prepends $\mathbf{k}$ to each input $\mathbf{x}_i$ and then obtains a privatized version $M(\llbracket\mathbf{k}; \mathbf{x}_i\rrbracket) = [\hat{k}_1, \ldots, \hat{k}_m, \hat{w}_1, \ldots, \hat{w}_n]$ of $\llbracket\mathbf{k}; \mathbf{x}_i\rrbracket$ through PCT2T. Let $\mathbf{z} = M(\llbracket\mathbf{k}; \mathbf{x}_i\rrbracket)$, with an example shown in Figure~\ref{fig:train_sample},  the LLM produces a sequence of activations $\bm{G}$ from the sequence of word embeddings $\bm{Z}$ of $\mathbf{z}$ together with the prompt $\bm{T}$:
\begin{align}
\bm{G} = f(\llbracket\bm{T}; \bm{Z}\rrbracket),
\end{align}
where $\bm{G} \in \mathbb{R}^{(l + m + n) \times h}$.

After obtaining $\bm{G}$, we use $m$ vectors $\bm{G}_{l:l+m} = [\bm{g}_{l+1}, \ldots, \bm{g}_{l+m}]$ in $\bm{G}$ to reconstruct the plain tokens. To achieve this, we introduce an additional reconstruction head consisting of two linear layers. The probability distribution for predicting the $i$-th token $k_i$ in the plain tokens is given by
\begin{align}
\bm{p}_i = \textrm{softmax}(\bm{W}_{\text{down}} \cdot \bm{W}_{\text{up}} \cdot \bm{g}_{l+i}),
\end{align}
where ``$\cdot$'' denotes matrix multiplication, $\bm{W}_{\text{down}} \in \mathbb{R}^{|V_r| \times c}$ and $\bm{W}_{\text{up}} \in \mathbb{R}^{c \times h}$ are the parameters of the reconstruction head, $c$ is the hidden size of the reconstruction head, and $|V_r|$ denotes the vocabulary size of the reconstruction head. The loss function for the privatized token reconstruction task is
\begin{align}
\mathcal{L}_{\textrm{rec}} = -\sum_{i=1}^{m} \log \bm{p}_i[j_i],
\end{align}
where $j_i$ denotes the index of $k_i$ in the reconstruction head vocabulary, and $\bm{p}_i[j_i]$ denotes the $j_i$-th scalar in the vector $\bm{p}_i$. Please note that the vocabulary of the reconstruction head does not need to be the same as the vocabulary of the LLM.

\begin{figure}[t]
    \centering
    \includegraphics[width=0.45\textwidth]{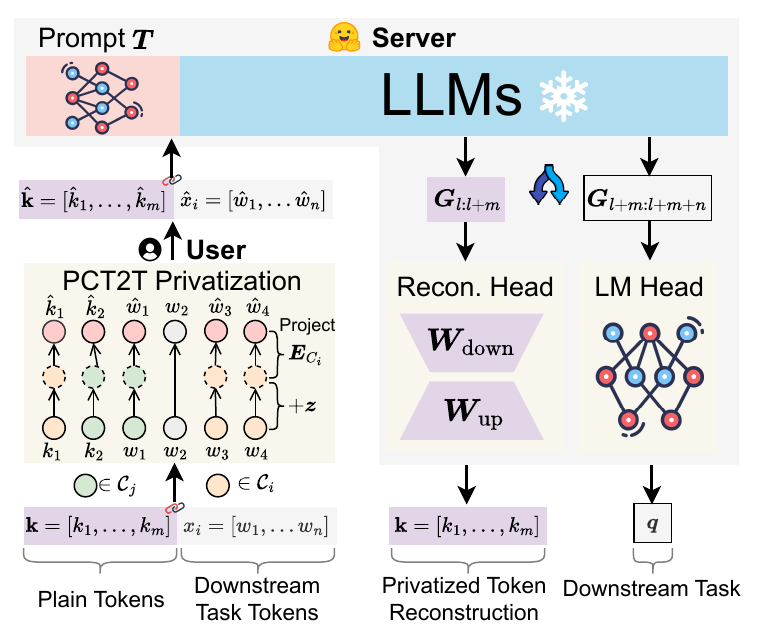}
    \caption{Overview of the \textsc{rapt} framework. \textsc{rapt} uses PCT2T to privatize data and trains with PEFT and token reconstruction for LLM customization.} 
    \label{fig:train}
\end{figure}

The remaining $n$ activations $\bm{G}_{l+m:l+m+n}$ are used to predict task-dependent labels. Assume the downstream task is a multi-class classification task, where $y_j \in \{0, 1\}$ is the label for the $j$-th class and $|C|$ is the number of classes.  The predicted probability distribution $\bm{q} \in \mathbb{R}^{|C|}$ is defined as
\begin{align}
\bm{q} & = \mathrm{softmax}(\frac{1}{n}\bm{W}_{\text{head}}\cdot\sum_{i=1}^n\bm{g}_{l+m+i}),
\end{align}
where $\bm{W}_{\text{head}} \in \mathbb{R}^{|C|\times h}$ denotes the parameters of the task-specific language modeling head. The objective function of the downstream task is formally described as
\begin{align}
\mathcal{L}_{\textrm{task}} = -\sum_{i=1}^{|C|} y_i \log \bm{q}[i].
\end{align}
As a result, the loss function for training \textsc{rapt} is given by $\mathcal{L} = \mathcal{L}_{\textrm{task}} + \mathcal{L}_{\textrm{rec}}$.

During the inference stage, the user also applies text-to-text privatization to protect private information and obtain predictions from the LLM service. It is worth noting that the reconstruction head is not used during inference. Therefore, $\{\bm{W}_{\text{down}}, \bm{W}_{\text{up}}\}$ can be discarded after training. Figure~\ref{fig:train} provides an illustration of \textsc{rapt} during LLM customization.

\section{Privacy Experiments}

We first evaluate the effectiveness of \textsc{rapt} against adversaries through simulated attacks, following the methodology of \citet{song2020information}, and conduct experiments on the UK section of the Trustpilot Sentiment dataset~(\textsc{TP-UK})~\cite{hovy2015user}. We use the embedding layers of \texttt{BERT$_\textrm{BASE}$} and \texttt{Qwen2.5-3B}~\cite{hui2024qwen2} to privatize texts in both T2T and PCT2T.

\subsection{Attacks and Metrics}

To evaluate the effectiveness of \textsc{rapt} in preserving privacy, we simulate the following types of attacks.

\subsubsection{Embedding Inversion Attack} This is a token-level attack in which the representations of user input text can be reversed to reveal the original input, potentially by an eavesdropping attacker. We consider the following types of attacks in this context~\cite{qu2021pcf}:

\paragraph{White-Box Attack} White-box attacks assume that attackers have full access to the model's weights, architecture, and training pipeline, allowing them to obtain gradient signals. However, we do not assume that attackers have access to the complete training data. Following the setting of \citet{qu2021pcf}, for any privatized token embedding $\bm{v}_{t}$ in the embedding space $\mathbb{R}^{d}$, the attacker aims to recover the original token embedding by solving $w_t = \argmin_{w_k} \lVert \bm{E}(w_k) - \bm{v}_{t} \rVert_{2}$. We employ nearest neighbor search to identify the original word for each perturbed word.

\paragraph{MLP-based Black-Box Attack} These attacks assume that attackers have access only to an API-like service, where they provide input and receive output without any additional information about the model. The attackers then use a neural network to predict the original data from its transformed representation.
    
\paragraph{Autoencoder-based Black-Box Attack} The attacker utilizes an autoencoder with an encoder $\phi$ and a decoder $\psi$. The encoder compresses the input $\bm{x}$ into a latent representation $\bm{z} = \phi(\bm{x})$, and then the decoder attempts to reconstruct the input as $\hat{\bm{x}} = \psi(\bm{z})$. The mean squared error (MSE) quantifies the reconstruction error, defined as $\text{MSE}(\bm{x}, \hat{\bm{x}}) = \|\bm{x} - \hat{\bm{x}}\|^2$. The autoencoder is trained to decode the privatized representations back to the original data, with a low MSE indicating a potential vulnerability.

\subsubsection{Attribute Inference Attack} In this attack, the attacker infers private attributes of a user from hidden representations of LLMs. Following \citet{plant2021cape}, let $\bm{z}_i$ denote the perturbed word embeddings of the LLM, and let $f_{\bm{\theta}}$ be a two-layer MLP with 768 hidden units and a ReLU activation function. The attacker predicts the distribution of private attributes as $\bm{p} = f_{\bm{\theta}}\left(\tfrac{1}{n}\sum_{i=1}^{n} \bm{z}_i \right)$, where $n$ is the sequence length. The loss function is defined as $\mathcal{L}_{\textrm{AIA}} = -\sum_{i=1}^{|C|} y_i \log \bm{p}[i]$, where $y_i \in \{0, 1\}$ is the label for the $i$-th private class and $|C|$ is the total number of private classes. We categorize age attributes in the TP-UK dataset into six equal-sized bins,\footnote{The bins are [$\leq$1955, 1955–1963, 1964–1971, 1972–1978, 1979–1985, $\geq$1986].} each assigned a unique label. The gender attribute in the TP-UK dataset is represented as a binary categorical variable.

\subsubsection{NER Attack} We assume the attacker employs a \texttt{BERT$_\textrm{BASE}$} model to perform Named Entity Recognition (NER) on input text sequences, extracting sensitive information such as names, addresses, and organizations~\cite{yang2023attack}. Given a sequence of input tokens $\mathbf{x} = [x_1, x_2, \ldots, x_n]$, the NER model assigns a probability distribution $\bm{p}_i$ to each token $x_i$, representing the likelihood of that token belonging to one of the predefined entity types. The attacker fine-tunes the \texttt{BERT$_\textrm{BASE}$} model by minimizing the objective function $\mathcal{L}_{\text{NER}} = -\sum_{i=1}^{n}\sum_{j=1}^{m} t_{ij} \log \bm{p}_{i}[j]$, where $t_{ij}$ is a binary indicator of whether token $x_i$ is of entity type $j$, and $\bm{p}_{i}[j]$ is the probability assigned by the model to that type.

For all types of attacks except autoencoder-based attacks, we use $1 - X$ as the evaluation metric, where $X$ denotes the attack success rate measured by accuracy or F1 score. For autoencoder-based attacks, we use MSE. We refer to these metrics collectively as \emph{empirical privacy}. Thus, a lower attack success rate corresponds to a higher value of empirical privacy, indicating stronger privacy protection.

\subsection{POS categories}

\begin{table}[t]
\centering
\caption{Example words in each selected POS category.}
\resizebox{0.45\textwidth}{!}{
\begin{tabular}{ll}
\toprule
\textbf{POS Category} & \textbf{Example Words}  \\\midrule
Noun & \emph{hospital, workplace, children, medication} \\ \midrule
Verb & \emph{visited, purchased, called, traveled} \\\midrule
Pronoun & \emph{I, my, his, her} \\\midrule
Preposition & \emph{at, from, to, near} \\\bottomrule
\end{tabular}}
\label{tab:pos_examples}
\end{table}

For the POS constraint, we carefully select \textit{Noun}, \textit{Verb}, \textit{Pronoun}, and \textit{Preposition} as the targets for privacy protection. \textit{Nouns} and \textit{Pronouns} have a direct connection to personally identifying information. \textit{Verbs} are included because they describe individual actions and behaviors, potentially exposing sensitive activity patterns. \textit{Prepositions} are also considered, as they often convey contextual details about direction or location that could inadvertently reveal personal information. We provide example words in these categories in Table~\ref{tab:pos_examples}.

However, the set of POS tags used varies considerably across languages. Consequently, the selected POS categories may differ between languages. Fortunately, certain POS categories, such as \emph{Noun} and \emph{Verb}, are universal and crucial for maintaining privacy, and thus can be included regardless of language. Other POS categories (\emph{e.g.}, postpositional markers in morphologically rich languages) should be selected by considering both language-specific characteristics and associated privacy implications. We defer a detailed study of the relationship between POS tags and privacy or utility in English to Section~\ref{sec:pos}. For highly specialized domains, such as legal or clinical texts, we found that sensitive information may span all POS categories. Therefore, in such cases, all categories should be considered. The key difference between PCT2T and the original T2T is that word substitution in PCT2T remains subject to POS constraints, whereas T2T does not impose such restrictions (\emph{e.g.}, a \emph{Noun} could be replaced with a \emph{Verb}).

\subsection{Results}

\begin{figure*}[!ht]
    \centering
    \includegraphics[width=\textwidth]{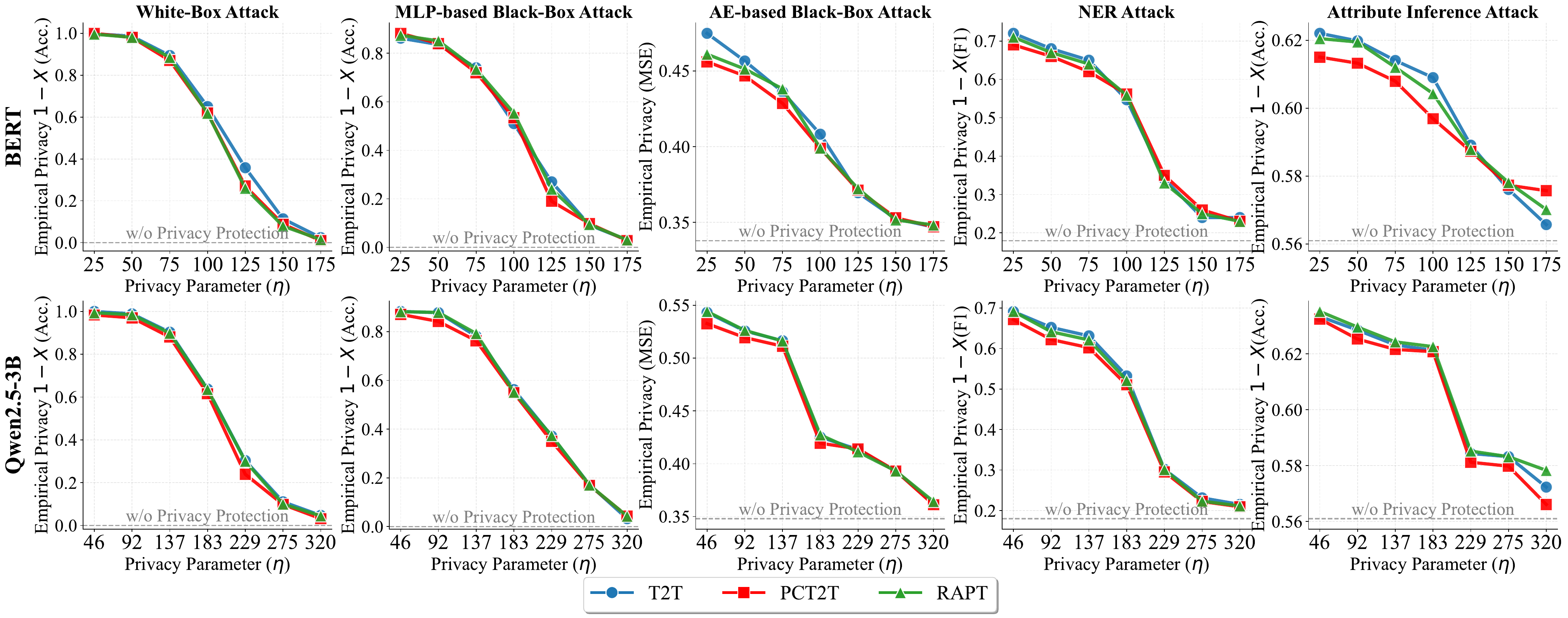}
    \caption{Results of the privacy experiments. The dashed horizontal lines indicate the baseline without privacy protection. ``AE'' denotes the autoencoder. Note that a smaller privacy parameter $\eta$ corresponds to stronger privacy protection.}

    \label{fig:privacy_eval}
\end{figure*}

Figure~\ref{fig:privacy_eval} illustrates the empirical privacy for T2T, PCT2T, and \textsc{rapt} (\emph{i.e.}, PCT2T + Reconstruction) under simulated attacks for \texttt{BERT$_\textrm{BASE}$} and \texttt{Qwen2.5-3B}. Because the average distance between two words in the embedding space differs across LLMs, we carefully adjust the privacy parameter $\eta$ for \texttt{Qwen2.5-3B} to make the results comparable with those of \texttt{BERT$_\textrm{BASE}$}. From the results, we make the following four observations. First, both T2T and PCT2T are effective in preserving privacy. Without privacy protection, empirical privacy remains low, indicating that attackers can easily reconstruct the original input. When applying T2T or PCT2T, empirical privacy increases substantially as the privacy parameter $\eta$ decreases. Second, T2T and PCT2T offer comparable privacy protection regardless of the value of $\eta$. Although adding an additional POS constraint may slightly weaken privacy protection compared to T2T, the effect is negligible when POS categories are carefully chosen. Third, adding a plain token reconstruction objective to PCT2T does not compromise privacy protection. This aligns with our intuition, since the original sentences are never reconstructed during training. Fourth, the empirical privacy results for \texttt{BERT$_\textrm{BASE}$} and \texttt{Qwen2.5-3B} are similar, indicating that privacy protection is achieved primarily through differential privacy rather than the choice of embedding model. Therefore, a smaller embedding model can be used locally without compromising privacy. We further investigate the relationship between utility and the embedding model in Section~\ref{sec:int_eval}. Therefore, we conclude that the \textsc{rapt} framework is effective in preserving privacy.

\section{Utility Experiments}
\begin{table*}[t]
\centering
\caption{Results for \textsc{prompt tuning}, \textsc{prefix-tuning}, \textsc{adapter}, and \textsc{fine-tuning} on the SST-2, QQP, and TP-UK tasks. Privacy parameters $\eta$ range from 25 to 175. The best results achieved with privacy preservation are highlighted in \textbf{bold}.
}\label{tab:complete_results}
\resizebox{\textwidth}{!}{
\begin{tabular}{lccccccccccccccccccc}
\toprule
\multirow{2}{*}{\textbf{Method}} & \multicolumn{6}{c}{\textbf{\textsc{SST-2}}} & \multicolumn{6}{c}{\textbf{\textsc{QQP}}} & \multicolumn{6}{c}{\textbf{\textsc{TP-UK}}} \\
\cmidrule(lr){2-7} \cmidrule(lr){8-13} \cmidrule(lr){14-19}
 &  25 &  75 & 125 & 150 & 175 & $+\infty$ &  25 & 75  & 125 & 150 & 175 & $+\infty$ &  25 & 75 & 125 & 150 & 175 & $+\infty$ \\\hline
\multicolumn{19}{c}{\texttt{BERT$_\textrm{BASE}$}} \\
\hline
\textsc{prompt tuning} \\
\quad$+$T2T Privatization& 46.3 & 47.1     & 55.6    & 56.3    & 68.8 &  88.3   & 51.2 & 55.4        & 63.6    & 69.8    & 73.4 & 87.7   & 55.4 & 67.1   & 79.3    & 81.9    & 83.6 & 88.7   \\
\quad$+$PCT2T Privatization& 51.8    &52.4    &60.5    &60.8    &72.9    &88.2    &56.1    &59.7    &68.2    &73.9    &78.7    & 87.8    &60.7    &71.5    &85.1    &85.9    &88.1    &88.5\\
\quad\quad$+$Reconstruction&\textbf{53.3}    &\textbf{53.6}    &\textbf{62.1}    &\textbf{63.1}    &\textbf{73.1}    & 88.5    &\textbf{56.5}    &\textbf{60.6}    &\textbf{69.1}    &\textbf{76.6}    &\textbf{80.2}    &87.4    &\textbf{62.2}    &\textbf{73.3}    &\textbf{86.0}    &\textbf{87.2}    & \textbf{88.5}   &88.6\\
\hdashline

\textsc{prefix-tuning} \\
\quad$+$T2T Privatization & 49.2    & 50.4    & 60.7    & 64.9    & 71.5 & 90.7    & 61.5    & 67.4    & 74.7    & 78.1    & 80.2  & 88.6  & 73.8    & 76.6    & 81.6    & 83.8    & 84.3  & 90.1  \\
\quad$+$PCT2T Privatization& 55.4	& 58.2 &	76.8 &	84.1 &	84.7 &	90.7&	66.0    &71.6    &78.9    &83.5    &85.4&	89.1&	77.5    &79.3    &83.9    &86.6    &88.8&	90.2\\
\quad\quad$+$Reconstruction&\textbf{56.4}&	\textbf{59.4}&	\textbf{78.9}&	\textbf{84.3}& \textbf{84.8}&	90.8&\textbf{67.6}    &\textbf{74.7}    &\textbf{81.9}    &\textbf{84.7}    &\textbf{85.9}&88.5&	\textbf{77.7}    &\textbf{79.7}    &\textbf{84.7}    &\textbf{88.1}    &\textbf{89.1}&	90.3
\\
\hdashline
\textsc{adapter (lora)}\\
\quad$+$T2T Privatization & 49.2    &53.1    &68.9    &76.9    &79.3 & 91.4   & 63.1    & 70.0    & 75.3    & 79.8    & 83.1  & 88.7  & 72.3    & 77.9    & 83.4    & 85.2    & 87.4  & 90.8  \\
\quad$+$PCT2T Privatization& 58.3    &60.6    &80.2    &88.1    &89.8&91.9&68.8    &75.3    &81.1    &85.4   &87.5&88.8&78.8&84.9&86.7&89.1&\textbf{90.5}&91.1\\
\quad\quad$+$Reconstruction&\textbf{59.7}&\textbf{62.4}&\textbf{81.1}&\textbf{89.6}&\textbf{89.9}&91.8&\textbf{69.2}    &\textbf{75.4}    &\textbf{82.9}    &\textbf{85.6}    &\textbf{87.6}&88.9&\textbf{81.0}&\textbf{85.3}&\textbf{88.4}&\textbf{89.9}&\textbf{90.5}&90.7\\
\hdashline
\textsc{fine-tuning} \\
\quad$+$T2T Privatization & 49.9    & 55.2    & 79.6    & 85.9    & 89.7 & 92.4  & 66.2    & 72.6    & 81.0    & 86.2    & 88.2   & 91.1 & 75.3    & 79.9    & 86.8    & 90.2    & 90.5 & 92.5   \\
\quad$+$PCT2T Privatization& 65.6    &70.4    &85.2&89.5&91.0&92.5&84.2&85.5&86.2&88.8&89.7&91.2&85.5&87.3&88.0&90.4&91.9&92.6\\
\quad\quad$+$Reconstruction&\textbf{66.0}    &\textbf{71.9}&\textbf{90.5}&\textbf{90.9}&\textbf{91.4}&92.8&\textbf{85.7}&\textbf{86.5}&\textbf{87.4}&\textbf{89.2}&\textbf{89.8}&91.6&\textbf{88.5}&\textbf{89.2}&\textbf{90.2}&\textbf{90.6}&\textbf{92.5}&92.7\\
 \hline
\multicolumn{19}{c}{\texttt{Qwen-2.5-3B}} \\
\hline
\textsc{prompt tuning} \\
\quad$+$T2T Privatization  &68.5    & 74.2    & 81.3    & 83.9    & 86.3  &  95.3 & 77.4    & 78.5    & 82.5    & 83.9    & 84.2  & 95.5  & 72.6    & 75.9    & 79.3    & 81.4    & 83.0   & 93.4 \\
\quad$+$PCT2T Privatization&75.3    &78.6    &88.3    &90.0    &91.6    &95.4    &82.1    &83.7    &85.6    &86.6    &87.1    &95.6    &79.6    &82.0    &85.7    &86.1    &87.0    &93.3\\
\quad\quad$+$Reconstruction&\textbf{76.6}    &\textbf{79.4}    &\textbf{90.9}    &\textbf{92.2}    &\textbf{93.3}    &95.4   &\textbf{84.5}    &\textbf{86.1}    &\textbf{89.6}    &\textbf{90.7}    &\textbf{91.5}    &95.4    &\textbf{82.8}    &\textbf{84.1}    &\textbf{87.1}    &\textbf{88.4}    &\textbf{89.5}    &93.3\\
\hdashline 
\textsc{prefix-tuning} \\
\quad$+$T2T Privatization  &70.0    &76.5    &83.8    &86.2    &89.2    &98.1    &85.1    &88.8    &88.4    &89.2    &91.5    &96.7    &80.1    &82.7    &84.9    &85.9    &86.2    &95.1  \\
\quad$+$PCT2T Privatization&79.3    &82.7    &91.4    &93.6    &94.4    &98.0    &88.3    &89.5    &90.2    &91.8    &93.2    &96.8    &82.9    &85.0    &88.5    &89.7    &90.0    &95.3\\
\quad\quad$+$Reconstruction&\textbf{81.2}    &\textbf{85.1}    &\textbf{93.8}    &\textbf{94.5}    &\textbf{95.7}    &97.9    &\textbf{89.4}    &\textbf{90.8}    &\textbf{92.5}    &\textbf{93.9}    &\textbf{94.3}    &96.9    &\textbf{84.7}    &\textbf{86.8}    &\textbf{89.9}    &\textbf{90.2}    &\textbf{90.9}    &95.2\\
\hdashline
\textsc{adapter (lora)}  \\
\quad$+$T2T Privatization  & 71.6    & 78.4    & 85.7    & 88.9    & 92.8 & 99.0   & 86.5    & 90.0    & 91.9    & 92.1    & 93.4  & 98.8  & 83.2    & 86.2    & 88.9    & 89.3    & 90.1  & 96.6  \\
\quad$+$PCT2T Privatization&80.4    &86.7    &94.4    &95.2    &96.8    &98.9    &89.1    &91.4    &93.4    &94.6    &95.7    &98.9    &86.5    &89.0    &92.4    &93.1    &93.7    &96.5\\
\quad\quad$+$Reconstruction&\textbf{84.9}    &\textbf{87.8}    &\textbf{96.6}    &\textbf{96.9}    &\textbf{97.3}    &98.7    &\textbf{91.3}    &\textbf{92.9}    &\textbf{95.6}    &\textbf{96.0}    &\textbf{96.6}    &98.6    &\textbf{90.5}    &\textbf{91.6}    &\textbf{93.5}    &\textbf{93.9}    &\textbf{94.7}    &96.5\\
\hdashline
\textsc{fine-tuning} \\
\quad$+$T2T Privatization  &81.6    & 88.8    & 93.3    & 94.8    & 95.3  & 99.5  & 92.7    & 93.8    & 95.8    & 96.3    & 97.2  & 99.6  & 90.4    & 91.7    & 93.7    & 94.6    & 95.2 & 98.4   \\
\quad$+$PCT2T Privatization&86.6    &90.5    &97.0    &97.9    &98.1    &99.6    &94.8    &95.7    &97.5    &98.1    &98.6    &99.7    &92.1    &93.3    &96.0   &96.2    &96.4    &98.3\\
\quad\quad$+$Reconstruction&\textbf{88.5}    &\textbf{92.5}    &\textbf{97.8}    &\textbf{98.8}    &\textbf{98.5}    &99.7    &\textbf{95.3}    &\textbf{96.5}    &\textbf{98.4}    &\textbf{98.7}    &\textbf{99.1}    &99.5    &\textbf{93.1}    &\textbf{94.2}    &\textbf{96.8}    &\textbf{97.0}    &\textbf{97.6}    &98.5\\
\bottomrule
\end{tabular}}
\label{tab:main_exp}
\end{table*}

\begin{table}[t]
    \centering
    \caption{Estimated word replacement probability for varying privacy parameters.}
    \resizebox{0.4\textwidth}{!}{
    \begin{tabular}{lccccc}
        \toprule
        \multirow{2}{*}{\textbf{Embedding}} &\multicolumn{5}{c}{\textbf{Word Replacement Probability}}\\
        \cmidrule(lr){2-6}
        & 25 & 75 &  125 & 150 & 175  \\
        \midrule
        \texttt{BERT}$_{\textrm{BASE}}$ & 0.94 & 0.85 & 0.45 & 0.14 & 0.05\\
  \bottomrule
    \end{tabular}}
    \label{tab:replacement_rate}
\end{table}

\subsection{Setup}
\subsubsection{Datasets, Metrics, and LLMs} We evaluate our approach on both Natural Language Understanding (NLU) and Natural Language Generation (NLG) tasks to demonstrate the versatility of \textsc{rapt} across different model architectures and application scenarios.  For NLU tasks, we use \texttt{BERT}\textsubscript{BASE}~\cite{devlin2018bert} and \texttt{Qwen2.5-3B}~\cite{hui2024qwen2} as backbone models and conduct experiments on three benchmark datasets: Stanford Sentiment Treebank (\textsc{SST-2})~\cite{socher2013recursive}, Quora Question Pairs (\textsc{QQP})~\cite{chen2018quora}, and \textsc{TP-UK}~\cite{hovy2015user}, with accuracy as the primary evaluation metric. The training/testing data sizes for SST-2, QQP, and TP-UK are 67,379/1,821, 363,846/390,965, and 48,647/6,080 examples, respectively. For NLG tasks, we adopt \texttt{Qwen2.5-3B} as the backbone LLM and evaluate on the \textsc{WebNLG} dataset~\cite{gardent2017webnlg}, which spans 14 domains and includes 18,102 training and 4,928 testing examples. We use BLEU as the evaluation metric and report results on the ``\textsc{All}'' test split. In these experiments, we consistently apply entity privatization to both input tables and generated texts, with privatized words restored on the user side after generation.

To assess long-context understanding, we use the \textsc{MMLU} benchmark~\cite{hendrycks2020measuring}. As \textsc{MMLU} is typically used for evaluation rather than training, we restrict our experiments to the ``\emph{auxiliary\_train}'' split, retaining 15,609 instances whose combined question and answer choice length does not exceed 500 tokens. We then apply a 70/30 split, yielding 10,926 training and 4,683 testing examples. This setup ensures consistency and meaningful comparison under privacy-preserving constraints.

\subsubsection{Baselines}
We build \textsc{rapt} on top of three representative PEFT methods. We also compare the \textsc{fine-tuning} method described in \citet{qu2021pcf}.
\begin{itemize}
\item \textbf{\textsc{prompt tuning}}~\cite{lester2021power}. A method that introduces learnable continuous virtual tokens for steering LLMs to downstream tasks.
\item \textbf{\textsc{prefix-tuning}}~\cite{li2021prefix}. This approach introduces more parameters by using deep prompts, which prepend continuous vectors to activations in all LLM layers.
\item \textbf{\textsc{adapter}} (\textsc{lora})~\cite{hu2021lora}. This method injects trainable low-rank matrices into every layer of the LLM.
\end{itemize}

\subsubsection{Implementation Details}
For NLU tasks, we use Adam~\cite{kingma2014adam} as the optimizer, with a learning rate of $6\times10^{-5}$ and a batch size of $128$. We train all tasks for $4$ epochs with a maximum input length of $128$ tokens. For \textsc{prompt-tuning}, we set the number of virtual tokens to $150$, while for \textsc{prefix-tuning}, the prefix length is set to $10$. We report results averaged over $5$ independent runs.

For NLG tasks, we employ AdamW~\cite{loshchilov2017decoupled} as the optimizer, with a learning rate of $2 \times 10^{-5}$, weight decay of $0.01$, and a batch size of $8$. We train all tasks for $4$ epochs with a maximum input length of $500$ tokens for \textsc{MMLU} and $128$ tokens for other tasks. For \textsc{prompt-tuning}, we set the number of virtual tokens to $100$, while for \textsc{prefix-tuning}, the prefix length is set to $30$. For \textsc{adapter} (\textsc{lora}), we set the hyperparameters as $r=16$, $\alpha=32$, and dropout $=0.05$. We employ beam search with a beam size of $5$, a repetition penalty of $1.2$, top-$k=4$, and top-$p=0.9$ for decoding.

All experiments are conducted on a single NVIDIA V100 GPU. For the privatized token reconstruction used in \textsc{rapt}, the hidden size $c$ is set to $96$, and the reconstruction head vocabulary size $|V_r|$ is set to $19,369$. Unless otherwise specified, we use the embedding layer of \texttt{BERT}\textsubscript{BASE} in PCT2T for all experiments. The number of plain tokens in \textsc{rapt} is set to $40$, sampled uniformly from selected POS categories. All other parameters are kept consistent with those used in the baseline methods for fair comparison.

\begin{table*}[t]
\centering
\caption{Results of \textsc{prompt tuning}, \textsc{prefix-tuning}, and \textsc{adapter} on the WebNLG and MMLU tasks with \texttt{Qwen2.5-3B}. \textbf{Bold} values indicate the best results under privacy protection.}
\resizebox{0.75\textwidth}{!}{
\begin{tabular}{lcccccccccc}
\toprule
\multirow{2}{*}{\hfil \textbf{Method}} & \multicolumn{5}{c}{\textbf{\textsc{WebNLG}}} & \multicolumn{5}{c}{\textbf{\textsc{MMLU}}} \\
\cmidrule(lr){2-6} \cmidrule(lr){7-11}
 & 25 & 75 & 125 & 175 & $+\infty$ & 25 & 75 & 125 & 175 & $+\infty$ \\
\hline
\textsc{prompt tuning} \\
\quad$+$T2T Privatization & 25.3 & 30.5 & 40.4 & 44.2 & 63.7 & 17.3 & 18.6 & 34.6 & 37.1 & 43.8 \\
\quad$+$PCT2T Privatization & 46.2 & 51.4 & 56.4 & 60.7 & 63.5 & 25.3 & 29.1 & 34.4 & 40.3 & 43.6 \\
\quad\quad$+$Reconstruction & \textbf{50.7} & \textbf{53.6} & \textbf{58.4} & \textbf{65.4} & 63.5 & \textbf{27.3} & \textbf{32.0} & \textbf{38.2} & \textbf{41.6} & 43.5 \\
\hdashline
\textsc{prefix-tuning} \\
\quad$+$T2T Privatization & 27.9 & 32.6 & 43.0 & 47.3 & 68.3 & 21.5 & 26.6 & 31.4 & 42.5 & 52.3 \\
\quad$+$PCT2T Privatization & 51.1 & 55.9 & 62.0 & 66.0 & 68.2 & 28.5 & 31.2 & 46.8 & 49.0 & 52.3 \\
\quad\quad$+$Reconstruction & \textbf{57.8} & \textbf{59.2} & \textbf{64.9} & \textbf{66.1} & 68.5 & \textbf{31.3} & \textbf{35.3} & \textbf{47.6} & \textbf{50.0} & 52.3 \\
\hdashline
\textsc{adapter (lora)} \\
\quad$+$T2T Privatization & 30.9 & 35.6 & 45.2 & 50.2 & 75.2 & 28.5 & 30.6 & 36.5 & 48.2 & 60.1 \\
\quad$+$PCT2T Privatization & 55.3 & 60.4 & 65.7 & 70.4 & 75.1 & 34.1 & 40.9 & 51.9 & 55.3 & 60.1 \\
\quad\quad$+$Reconstruction & \textbf{63.7} & \textbf{63.9} & \textbf{67.1} & \textbf{73.3} & 75.1 & \textbf{35.7} & \textbf{42.2} & \textbf{53.2} & \textbf{56.3} & 60.1 \\
\hdashline
\textsc{fine-tuning} \\
\quad$+$T2T Privatization & 56.4 & 63.1 & 67.7 & 72.6 & 88.1 & 55.1 & 59.2 & 68.4 & 71.0 & 78.4 \\
\quad$+$PCT2T Privatization & 74.5 & 76.2 & 83.1 & 84.4 & 88.1 & 60.4 & 64.6 & 71.1 & 74.3 & 78.5 \\
\quad\quad$+$Reconstruction & \textbf{76.0} & \textbf{78.0} & \textbf{84.2} & \textbf{86.3} & 88.1 & \textbf{61.4} & \textbf{65.9} & \textbf{72.9} & \textbf{75.6} & 78.4 \\
\bottomrule
\end{tabular}}
\label{tab:main_exp_llms}
\end{table*}

\subsection{Results on NLU tasks}

Table~\ref{tab:main_exp} shows the results of \textsc{prompt tuning}, \textsc{prefix-tuning}, \textsc{adapter}, and \textsc{fine-tuning} with and without privacy protection on three NLU tasks. We give word replacement probabilities for different privacy parameter $\eta$ in Table~\ref{tab:replacement_rate}, which serves as an intuitive measure for comparing the degree of privacy protection between different $\eta$. Note that our proposed \textsc{rapt} approach \textit{incorporates both PCT2T privatization and privatized token reconstruction}.  The columns with $\eta=+\infty$ represent results without privacy protection. 

From Table~\ref{tab:main_exp}, we have the following observations: First, with smaller $\eta$, text-to-text privatization (both T2T and PCT2T) offers stronger privacy guarantees but it inevitably hurts the performance of downstream tasks. Second, when adopting T2T privatization, we can see that both \textsc{prompt tuning} and \textsc{prefix-tuning} perform poorly across all tasks and LLMs even with the weakest privacy protection (the largest $\eta$), which indicates that prompt tuning methods are sensitive to random perturbations imposed by text-to-text privatization. Third, using PCT2T privatization instead of T2T privatization consistently improves the performance across different choices of $\eta$, which indicates the effectiveness of PCT2T in preserving the syntactic and semantics of the original sentence. Fourth, by further introducing the privatized token reconstruction task, the performance of all four methods is significantly improved across tasks and LLMs. Adding token reconstruction cannot bring further improvements when $\eta=+\infty$. This is because reconstructing plain tokens becomes trivial in the absence of privacy protection.

These results suggest that both PCT2T and privatized token reconstruction are very effective in improving the performance of PEFT methods when trained on privatized data. The results also coincide with our intuition that using reconstruction can help LLMs learn better representations.

\subsection{Results on NLG tasks}

To validate the effectiveness of our method on NLG tasks, we conduct additional experiments on the WebNLG and MMLU datasets using \texttt{Qwen2.5-3B}, with the results presented in Table~\ref{tab:main_exp_llms}. We observe that performance drops significantly when T2T privatization is applied, consistent with the results for NLU tasks. In contrast, PCT2T achieves substantially better performance than T2T, indicating its effectiveness in preserving the syntax and semantics of the inputs. When combined with privatized token reconstruction, performance improves further and significantly. These results confirm the versatility and effectiveness of our proposed PCT2T and privatized token reconstruction methods in preserving privacy.

\subsection{Intrinsic Evaluation}\label{sec:int_eval}
We study the characteristics of \textsc{rapt} by comparing different variants of \textsc{rapt} using \texttt{BERT$_\textrm{BASE}$} and \texttt{Qwen2.5-3B} model. We assume that \textsc{rapt} utilizes \textsc{prefix-tuning} to steer \texttt{BERT$_\textrm{BASE}$} on the SST-2 task.

\subsubsection{Embedding Models}
\begin{table}[t]
\centering
\caption{Results on the SST-2 task using different embedding models for PCT2T. ``Pr.'' denotes the probability of replacing a word with PCT2T privatization. ``ACC.'' denotes accuracy.}
\resizebox{0.45\textwidth}{!}{
\begin{tabular}{l@{\hskip 0.5in}r@{\hskip 0.5in}r@{\hskip 0.5in}r}
\toprule
\textbf{Embedding} & \textbf{Pr.} & $\mathbf{\eta}$ & \textbf{ACC.} \\\midrule
BERT & 0.14 & 150 & 84.3 \\ 
BioBERT & 0.16  & 125 & 83.6 \\
GPT-2 & 0.15 & 60 & 83.1 \\
RoBERTa & 0.14 & 50 & 85.7 \\
T5 & 0.12 & 5 & 87.3 \\\bottomrule
\end{tabular}}
\label{tab:exp_emb}
\end{table}

\begin{table}[t]
\centering
\caption{Results of \textsc{prefix-tuning} on SST-2 with varying prompt length and $\eta$. ``Params.'' denotes the number of parameters in the trainable prompt.}
\resizebox{0.45\textwidth}{!}{
\begin{tabular}{rcccc}
\toprule
\textbf{Length} & \textbf{Params.} & \textbf{$\eta = 75$} & \textbf{$\eta = 125$} & \textbf{$\eta = 150$}  \\
\midrule
10  & 0.61M & 59.4 & 78.9 & 84.3 \\
40  & 0.95M &  60.4 & 79.5 & 85.6 \\
100 & 1.51M & 61.8 & 80.6 & 87.5 \\
\bottomrule       
\end{tabular}
}
\label{tab:prompt_tuning}
\end{table}

\begin{table}[t]
\centering
\caption{Results of different plain token selection methods.}
\label{tab:noise_scale_comparison}
\resizebox{0.45\textwidth}{!}{
\begin{tabular}{lcccc}
\toprule
\multirow{2}{*}{\textbf{Method}} & \multicolumn{2}{c}{\textbf{SST-2}} & \multicolumn{2}{c}{\textbf{WebNLG}} \\
\cmidrule(lr){2-3} \cmidrule(lr){4-5}
 & $\eta=75$ & $\eta=125$ & $\eta=75$ & $\eta=125$ \\
\midrule
Uniform & 85.1 & 93.8 & 59.2 & 64.9 \\
Frequency & 85.2 & 94.5 & 60.3 & 65.2 \\
\bottomrule
\end{tabular}}
\end{table}

We investigate the effect of using different embedding models in PCT2T privatization. Specifically, the user side uses embedding models from GPT-2, RoBERTa, BioBERT~\cite{lee2020biobert}, or T5 to map the input text to the privatized text. We adjust the privacy parameter $\eta$ for different embedding models to match the level of probability for replacing a token in the text. Table~\ref{tab:exp_emb} shows the results of our experiments. We find that using different embedding models does not significantly affect performance on the downstream task, even with embedding models trained on different domains (\emph{e.g.}, BERT vs. BioBERT).

\subsubsection{Prompt Length}
Table~\ref{tab:prompt_tuning} shows the performance of \textsc{rapt} using different prompt lengths. We observe that \textsc{rapt} generally performs better as the prompt length increases. These results also align with previous studies~\cite{li2021prefix,lester2021power}.

\subsubsection{Privatized Token Reconstruction}
We study the effect of privatized token reconstruction by varying the content of plain tokens, the number of plain token examples used for training, the length of plain tokens, and the hidden size of the reconstruction head.
\paragraph{Content of Plain Tokens} We first demonstrate that the content of plain tokens can be chosen arbitrarily. To verify this, we randomly generate five plain token sequences of length 40 and evaluate their performance with \texttt{BERT$_\textrm{BASE}$} on the SST-2 task. The results range from 84.2 to 84.7, indicating that different token choices yield comparable performance.  We then study different methods for selecting plain tokens. Specifically, we compare frequency-based sampling of plain tokens, guided by the POS tag and token distributions in the training data, with uniform sampling. The results, shown in Table~\ref{tab:noise_scale_comparison}, indicate that the frequency-based approach slightly improves downstream task performance, suggesting that prior knowledge of the training data can enhance utility. However, this raises privacy concerns, as frequency-based replacements may be exploited by adversaries to infer the data domain or sensitive attributes. Therefore, we use random selection to maintain a balanced trade-off between privacy and utility.

\begin{figure}[t]
    \centering
    \subfloat[]{\includegraphics[width=0.163\textwidth]{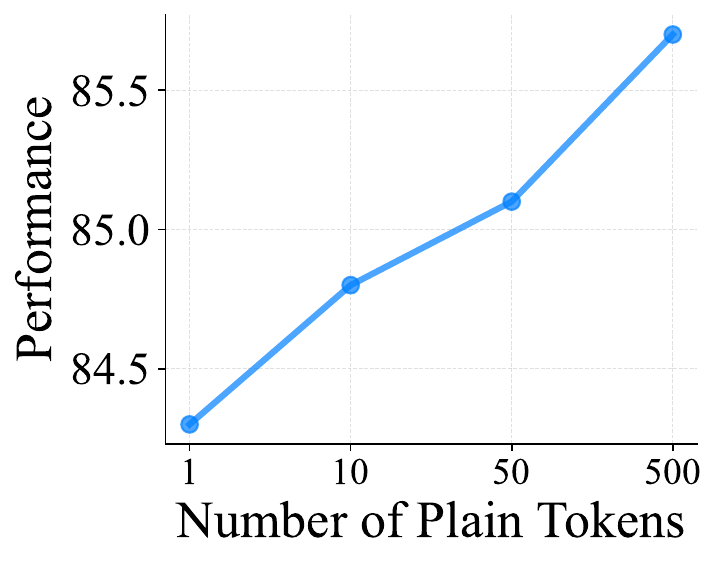}}%
    \subfloat[]{\includegraphics[width=0.163\textwidth]{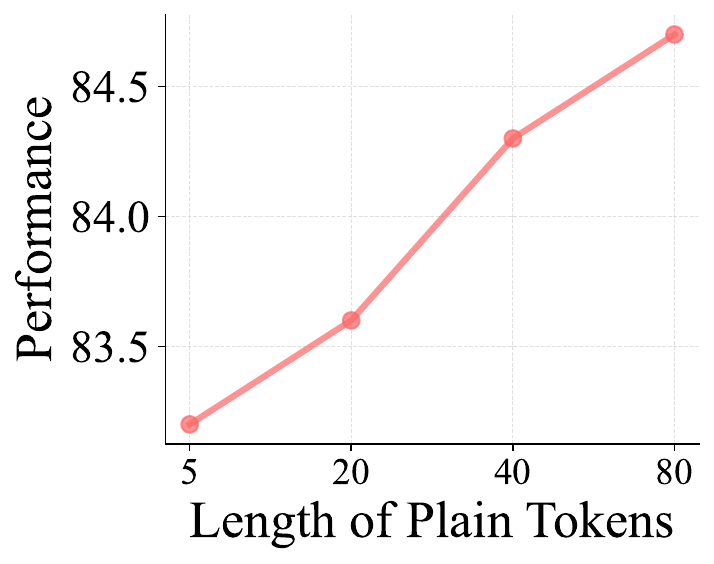}}%
    \subfloat[]{\includegraphics[width=0.163\textwidth]{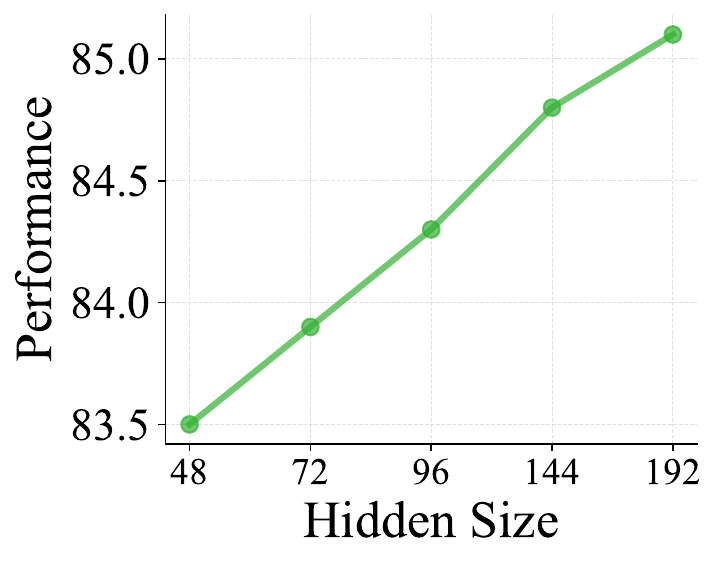}}%
    \caption{The effects of the number of plain tokens, the length of plain tokens, and the hidden size of the reconstruction head on the SST-2 task.}
    \label{fig:plain_token_effect}
\end{figure}

\begin{table}[t]
\centering
\caption{Comparison of \textsc{rapt} with plain tokens for token reconstruction, without the reconstruction objective, and without plain tokens entirely.}
\label{tab:plain_effect_compare}
\resizebox{0.45\textwidth}{!}{
\begin{tabular}{lcccc}
\toprule
\multicolumn{1}{c}{\multirow{2}{*}{\textbf{Method}}} & \multicolumn{2}{c}{\textbf{SST-2}} & \multicolumn{2}{c}{\textbf{WebNLG}} \\
\cmidrule(lr){2-3} \cmidrule(lr){4-5}
 & \textsc{$\eta=75$} & \textsc{$\eta=125$} & \textsc{$\eta=75$} & \textsc{$\eta=125$} \\
\midrule
\textsc{rapt} & \textbf{85.1} & \textbf{93.8} & \textbf{59.2} & \textbf{64.9} \\
\emph{w/o} Reconstruction & 80.8 & 90.1 & 53.2 & 60.0 \\
\emph{w/o} Plain Tokens & 82.7 & 92.5 & 55.9 & 62.0 \\
\bottomrule
\end{tabular}}
\end{table}

\begin{figure*}[t]
    \centering
    
    \subfloat[]{\includegraphics[width=0.25\textwidth]{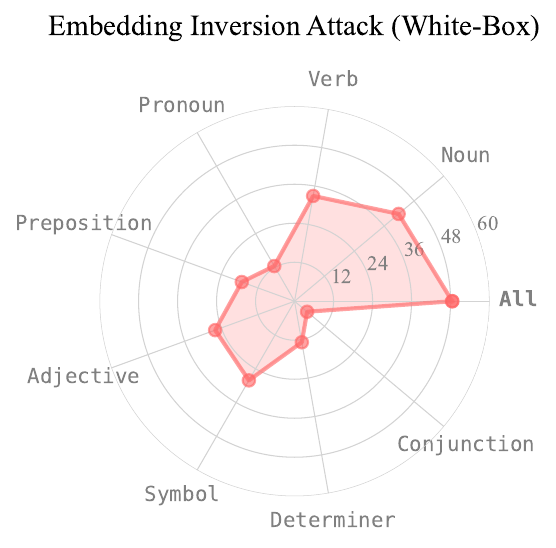}}%
    \subfloat[]{\includegraphics[width=0.25\textwidth]{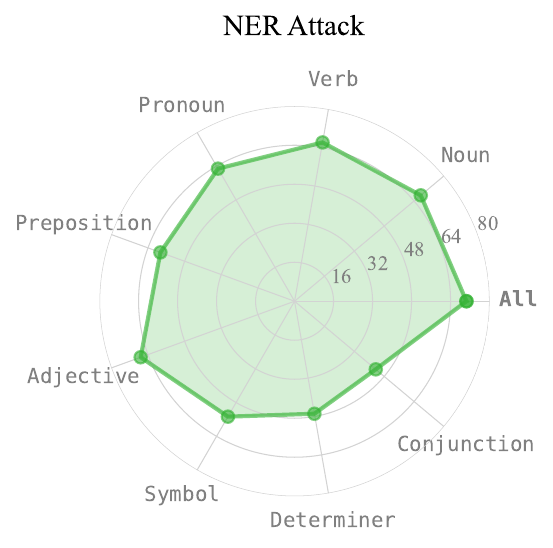}}%
    \subfloat[]{\includegraphics[width=0.25\textwidth]{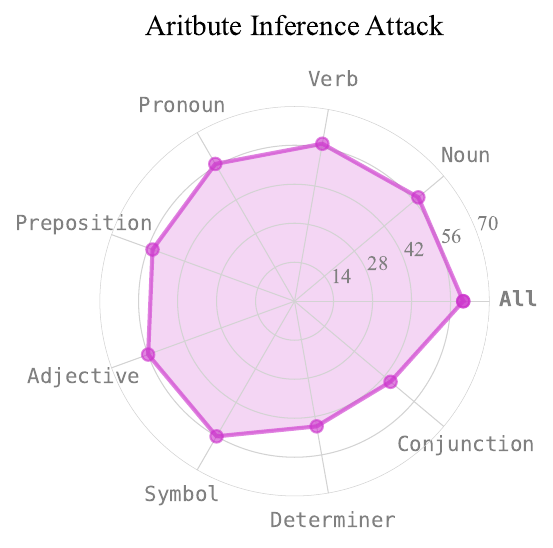}}
    \subfloat[]{\includegraphics[width=0.25\textwidth]{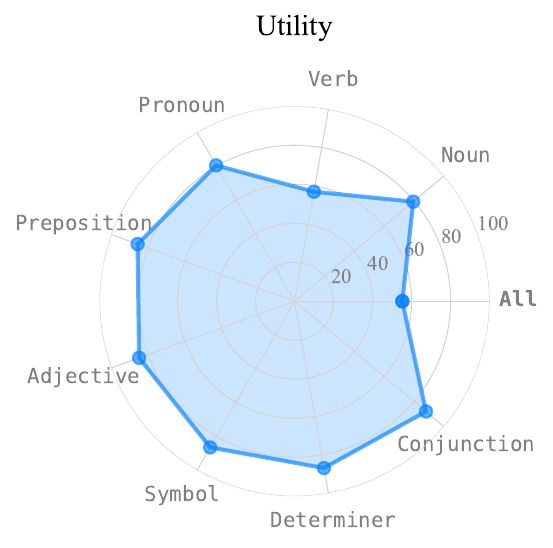}}%

    \caption{The effect of using different POS categories on privacy and utility. We simulate the embedding inversion attack, NER attack, and attribute inference attack with $\eta = 100$. The ``All'' category represents selecting all POS tags for privacy protection.}
    \label{fig:POS_analysis}
\end{figure*}

\paragraph{Number of Plain Tokens} We study the effect of using diverse plain tokens in the privatized token reconstruction task. We first construct a set of plain tokens with the same length. Then, during training, for each training input, we randomly choose plain tokens and prepend them to the input. From Figure~\ref{fig:plain_token_effect}~(a), we find that using more plain tokens during training slightly improves performance on SST-2. However, using one plain token during privatized token reconstruction suffices to improve the performance of the LLM on downstream tasks.

\paragraph{Length of Plain Tokens} We study the effect of using plain tokens with different lengths. Intuitively, the LLM needs to learn better representations to reconstruct longer plain tokens. Therefore, using longer plain tokens may benefit the corresponding performance on downstream tasks. As shown in Figure~\ref{fig:plain_token_effect}~(b), we find that the results match our intuition: using longer plain tokens performs significantly better than using shorter plain tokens. To verify that the performance improvement on downstream tasks does not result from the addition of plain tokens, we conduct an ablation study on \texttt{Qwen2.5-3B} by incorporating plain tokens during \textsc{prefix-tuning} without introducing the reconstruction objective. As shown in Table~\ref{tab:plain_effect_compare}, adding plain tokens during training without the reconstruction objective performs significantly worse than \textsc{rapt}, and even underperforms the variant that does not include plain tokens (equivalent to ``$+$Privatization'' in Table~\ref{tab:main_exp_llms}). These results confirm that the plain token reconstruction is essential for achieving improved performance.

\paragraph{Reconstruction Head}
We study the hidden size and vocabulary size of the reconstruction head. Figure~\ref{fig:plain_token_effect}~(c) shows the results. Using a larger hidden size generally performs better but introduces more parameters during training. As for the vocabulary size, we found it is necessary to use a moderately large vocabulary. A small vocabulary makes the prediction of plain tokens much easier, therefore degrading the benefits of privatized token reconstruction task.

To summarize, we found that \textsc{rapt} is robust to hyperparameter choices. Using a longer prompt length, larger hidden and vocabulary sizes for the reconstruction head, and diverse and longer plain tokens can generally improve performance on downstream tasks, at the cost of increased computational and memory overhead. A moderate choice of these hyperparameters is sufficient to achieve significant improvements over the baselines while maintaining relatively low overhead.

\subsection{Effect of POS Categories}\label{sec:pos}

We investigate the effect of choices of POS categories on both utility and privacy for PCT2T. We focus on embedding inversion attacks, NER attacks, and attribute inference attacks. The privacy parameter $\eta$ is set to 100. All utility experiments are conducted on the SST-2 task. For POS categories, we investigate using \textit{Noun}, \textit{Verb}, \textit{Pronoun}, \textit{Preposition}, \textit{Adjective}, \textit{Symbol}, \textit{Determiner}, and \textit{Conjunction}. Figure~\ref{fig:POS_analysis} presents the results, from which we derive several key findings. First, POS categories such as \textit{Conjunctions} and \textit{Determiners} provide relatively weaker privacy protection, indicating that not all POS categories are equally useful for privacy preservation. However, these categories are important for maintaining the syntactic structure of the input sentence. This finding supports our motivation for the PCT2T privatization approach. Second, POS categories like \textit{Noun} and \textit{Verb} are effective for privacy preservation. These categories are closely related to personal identification information. Therefore, we include \textit{Noun} and \textit{Verb} in our set of POS categories for all our experiments. 

\subsection{Comparison with Other Methods}

\begin{figure}[t]
    \centering
    \includegraphics[width=0.49\textwidth]{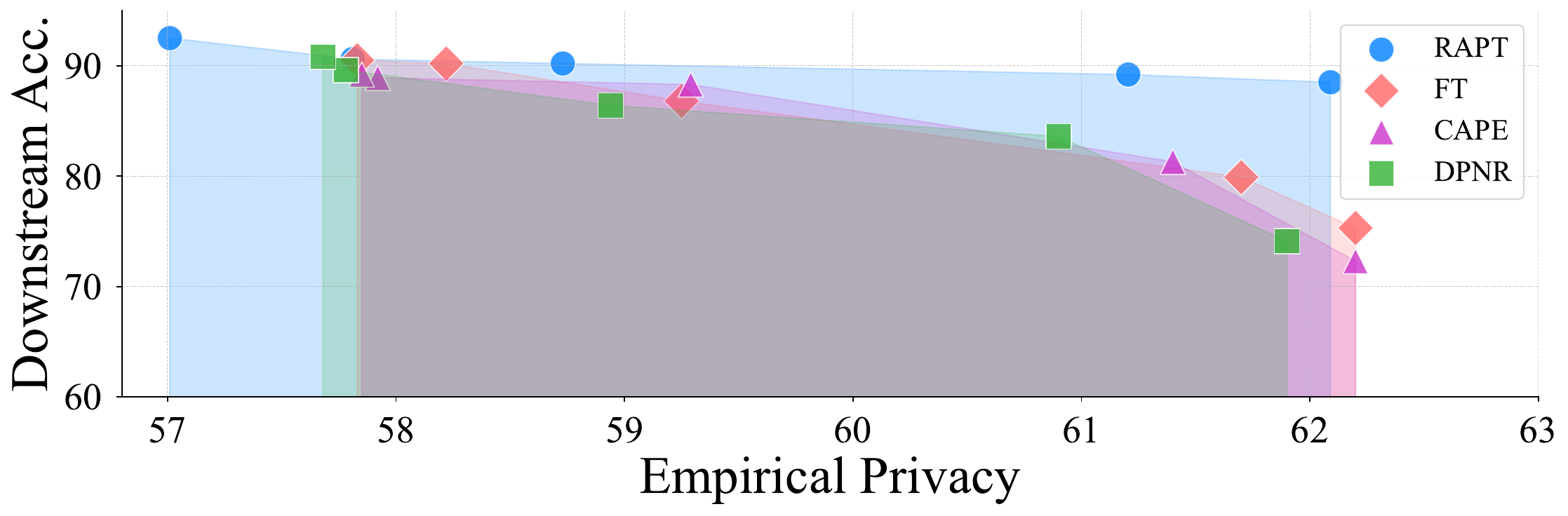}
    \caption{Pareto frontiers for \textsc{rapt}, \textsc{fine-tuning}, CAPE, and DPNR. We simulate an attribute inference attack to calculate empirical privacy. The dataset used for the downstream task is the TP-UK dataset.}
    \label{fig:Pareto_Frontiers_baselines}
\end{figure}

We compare \textsc{rapt} with other methods in terms of both privacy and utility. We consider the following methods:
\begin{itemize}
\item \textbf{CAPE}~\cite{plant2021cape}. This approach uses a Laplace-based embedding perturbation coupled with an adversarial learning objective to protect privacy. The method privatizes the output representations of LLMs.
\item \textbf{DPNR}~\cite{lyu2020differentially}. The method introduces a combination of word dropout and Laplace-based embedding perturbation. It also privatizes the output representations of LLMs.
\item \textbf{Fine-Tuning}~\cite{qu2021pcf}. This method privatizes the embedding layer of LLMs. It applies $d_{\mathcal{X}}$-privacy to perturb input embeddings of the user.
\end{itemize}

To ensure a fair comparison between these methods, we use the commonly used attribute inference attack to study the privacy-utility trade-off of different methods. For CAPE, we utilize the \texttt{BERT$_\textrm{BASE}$} model with privacy parameters set as $\epsilon = 0.01$ and $\lambda = 1.5$, and the adversarial training learning rate is set to $1\times10^{-3}$. For DPNR, we also use the \texttt{BERT$_\textrm{BASE}$} model. The word dropout rate $\mu$ is set to $0.1$ and the privacy parameter $\epsilon$ is set to $5$. For our implementation of \textsc{rapt}, we set the privacy parameter $\eta$ to $100$.

Figure~\ref{fig:Pareto_Frontiers_baselines} illustrates the Pareto Frontiers for different methods. We can observe that \textsc{rapt} consistently achieves better utility than CAPE, DPNR, and \textsc{fine-tuning} under the same degree of privacy protection. As the degree of empirical privacy increases, the performance gap between \textsc{rapt} and the other methods becomes larger. The results confirm that our \textsc{rapt} method is effective in preserving the privacy of LLMs.

\subsection{Extended Evaluation on Additional LLMs}
\begin{table}[t]
\centering
\caption{Results of Llama3-8B and Mistral-7B on the SST-2 task.}
\resizebox{0.45\textwidth}{!}{
\begin{tabular}{lcccccc}
\toprule
\textbf{Method} &  25 &  75 & 125 & 150 & 175 & $+\infty$ \\
\midrule
\texttt{Llama3-8B} \\
\quad$+$T2T Privatization& 69.8 & 75.6     & 78.6    & 85.5    & 89.6 &  96.6  \\
\quad$+$PCT2T Privatization& 76.8 & 79.4     & 83.6    & 87.2    & 93.6  &  96.4  \\
\quad\quad$+$Reconstruction&\textbf{77.6} & \textbf{80.3}     & \textbf{85.2}    & \textbf{88.4}    & \textbf{93.7} &  96.5\\
\midrule
\texttt{Mistral-7B} \\
\quad$+$T2T Privatization& 65.3 & 72.6     & 75.1    & 83.4    & 87.2 &  94.1  \\
\quad$+$PCT2T Privatization& 68.9 & 73.5     & 75.4    & 85.6    & 90.2 &  94.3  \\
\quad\quad$+$Reconstruction&\textbf{70.1} & \textbf{75.1}     & \textbf{76.2}    & \textbf{86.4}    & \textbf{91.5} &  94.2\\
\bottomrule
\end{tabular}}
\label{tab:LLMs_complete_results}
\end{table}

To validate the effectiveness of our method on recent
LLMs, we further conducted experiments on the SST-2 task using \texttt{Llama3-8B}~\cite{dubey2024llama} and \texttt{Mistral-7B}~\cite{jiang2023mistral7b}. We utilized the \texttt{BERT$_\textrm{BASE}$} embeddings from the model for applying T2T and PCT2T
and fine-tuning the LLMs with the QLoRA~\cite{dettmers2024qlora} method. Table~\ref{tab:LLMs_complete_results} shows the results. From the results, we made the
same observations as in previous experiments, indicating that
our methods are applicable to different LLMs. As a result, we conclude that \textsc{rapt} is effective in achieving better utility for
the preservation of LLM privacy.

\section{Analyses}

\subsection{Embedding Distribution and Privacy Preservation}
\begin{figure}[t]
    \centering
    \includegraphics[width=0.42\textwidth]{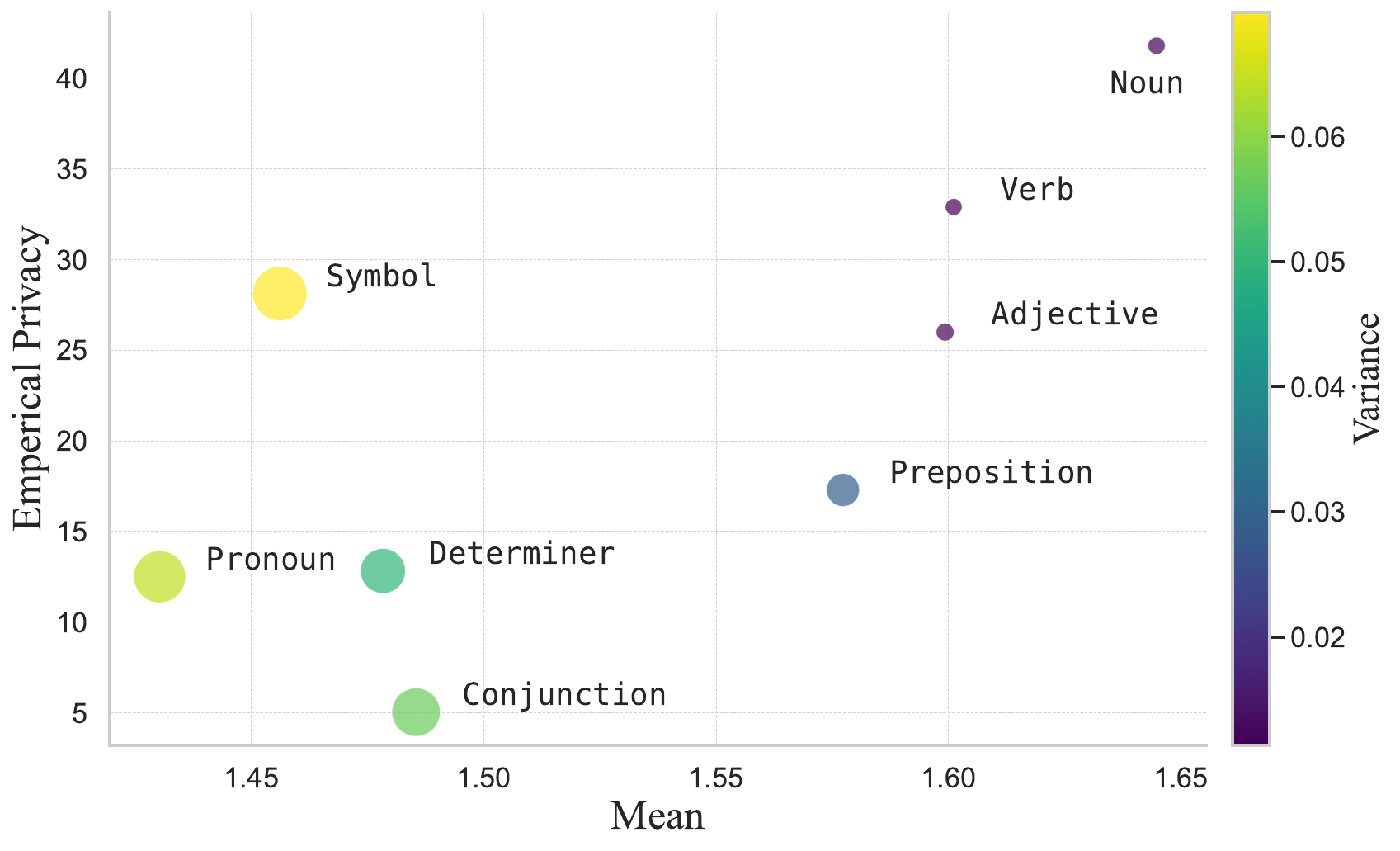}
    \caption{Embedding distribution across different POS categories and their corresponding empirical privacy. The $x$-axis represents the mean distance of each category, while the $y$-axis indicates the degree of empirical privacy. The color of each circle corresponds to the variance in distance for each POS category.}
    \label{fig:pos_distribution}
\end{figure}

We first investigate the distribution of word embeddings within a specific POS category and the corresponding empirical privacy using PCT2T. For each pair of words $w_i$ and $w_j$ in a POS category, we compute the pairwise Euclidean distance as $d_{ij} = | \bm{w}_i - \bm{w}_j |$. Then, the mean distance is calculated as $\mu = \frac{2}{n(n - 1)} \sum_{i=1}^{n-1} \sum_{j=i+1}^{n} d_{ij}$, where $n$ is the total number of words. The variance of the distance is calculated as $\sigma^2 = \frac{2}{n(n - 1)} \sum_{i=1}^{n-1} \sum_{j=i+1}^{n} (d_{ij} - \mu)^2$.

We plot the mean and variance of the distances and the corresponding empirical privacy for a given POS category in Figure~\ref{fig:pos_distribution}. From the figure, we can see that the categories \textit{Noun} and \textit{Verb} achieve better privacy protection compared to other categories. We also find that the mean distance and variance for the \textit{Noun}, \textit{Verb}, and \textit{Adjective} categories are similar. This finding reveals that it is necessary to differentiate between different POS categories in text-to-text privatization.

\subsection{Geometry of Representations}

\begin{figure}[t]
    \centering
     \includegraphics[width=0.45\textwidth]{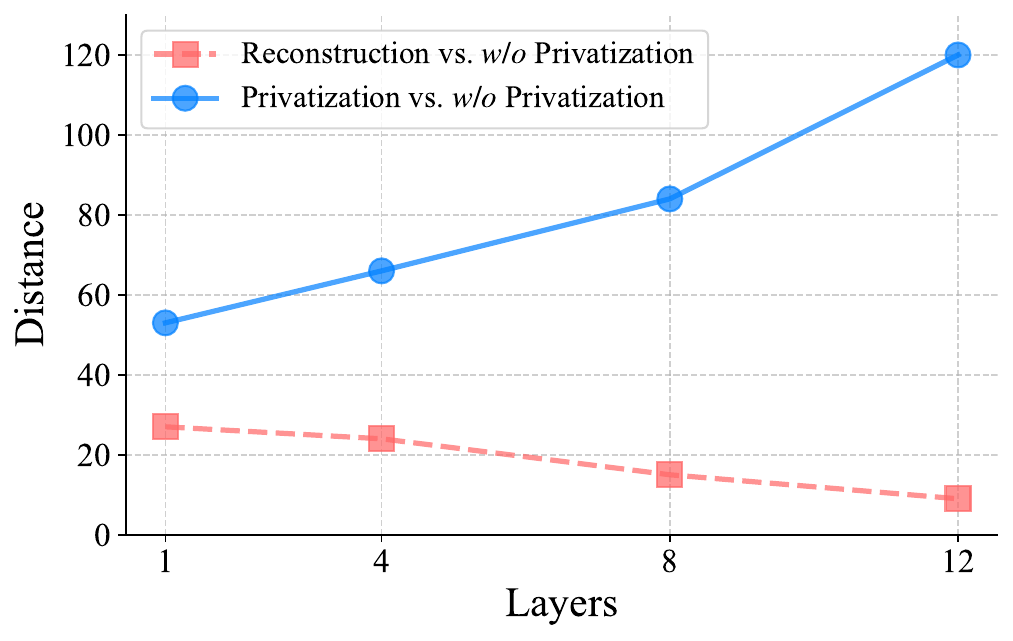}
     \caption{We calculate the average distance in different layers of representations between different methods.}
     \label{fig:distance}
\end{figure}

We analyze the geometry of LLM representations to investigate why \textsc{rapt} enhances the performance of downstream tasks when trained on locally privatized data. Specifically, we examine \texttt{BERT$_\textrm{BASE}$} customized with \textsc{rapt}, using \textsc{prefix tuning} and setting $\eta=100$. We first compute representations at each BERT layer for \textsc{prefix-tuning} without privatization, \textsc{prefix-tuning} with T2T privatization, and \textsc{rapt}, using data from SST-2. For each layer, we then project the representations into 2D space using Principal Component Analysis (PCA) and compute the average distance between representations of the different methods. Figure~\ref{fig:distance} shows the results. We found that the deep representations learned by \textsc{rapt} are similar to those learned without privacy protection. Notably, the representations for \textsc{rapt} become progressively closer to those without privacy protection as the layer number increases. These results confirm that \textsc{rapt} can learn better representations.

\subsection{Pareto Frontiers}

\begin{figure}[t]
    \centering
    \subfloat[]{\includegraphics[width=0.25\textwidth]{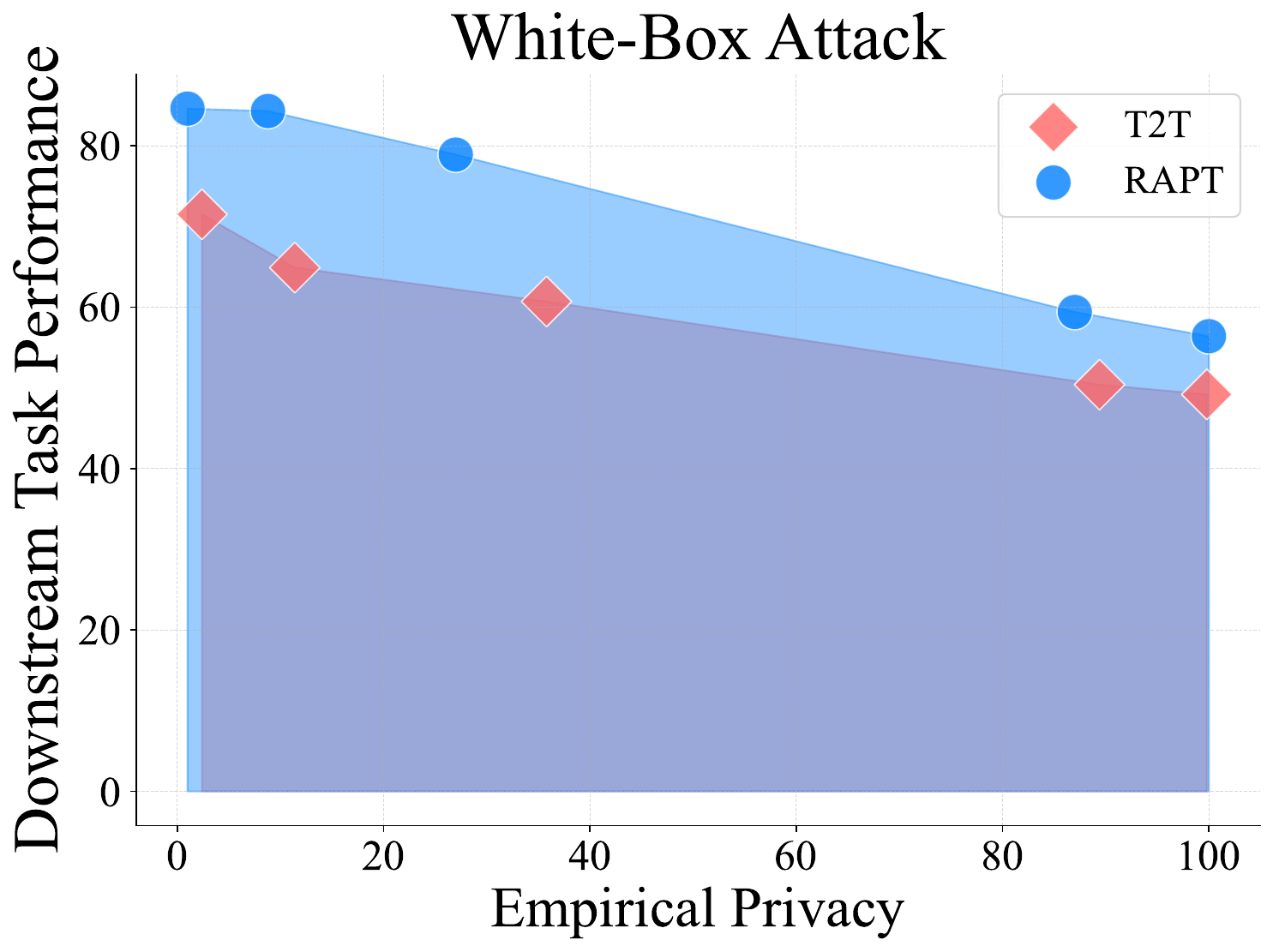}}%
    \subfloat[]{\includegraphics[width=0.25\textwidth]{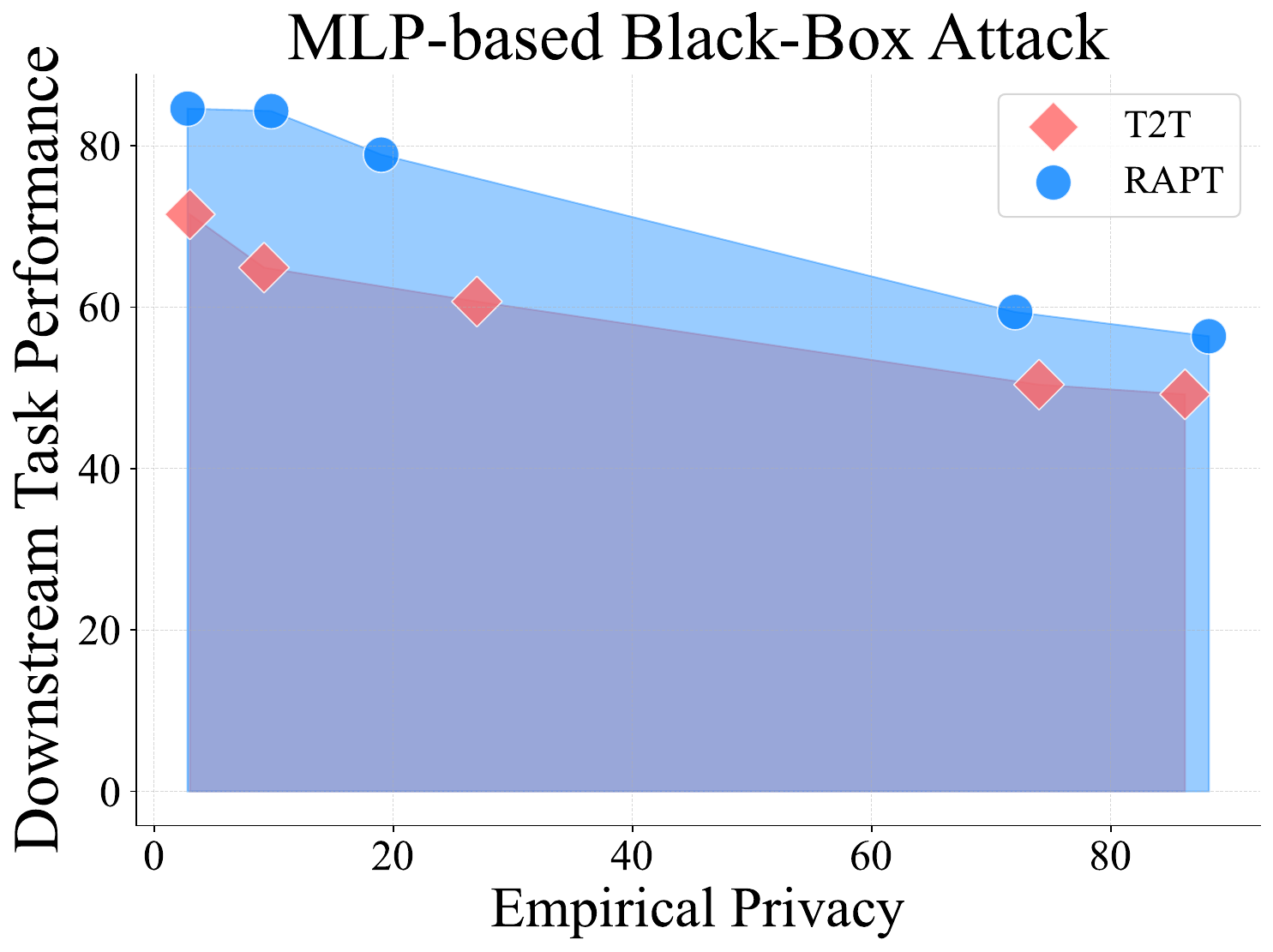}}\\[5pt]
    \subfloat[]{\includegraphics[width=0.25\textwidth]{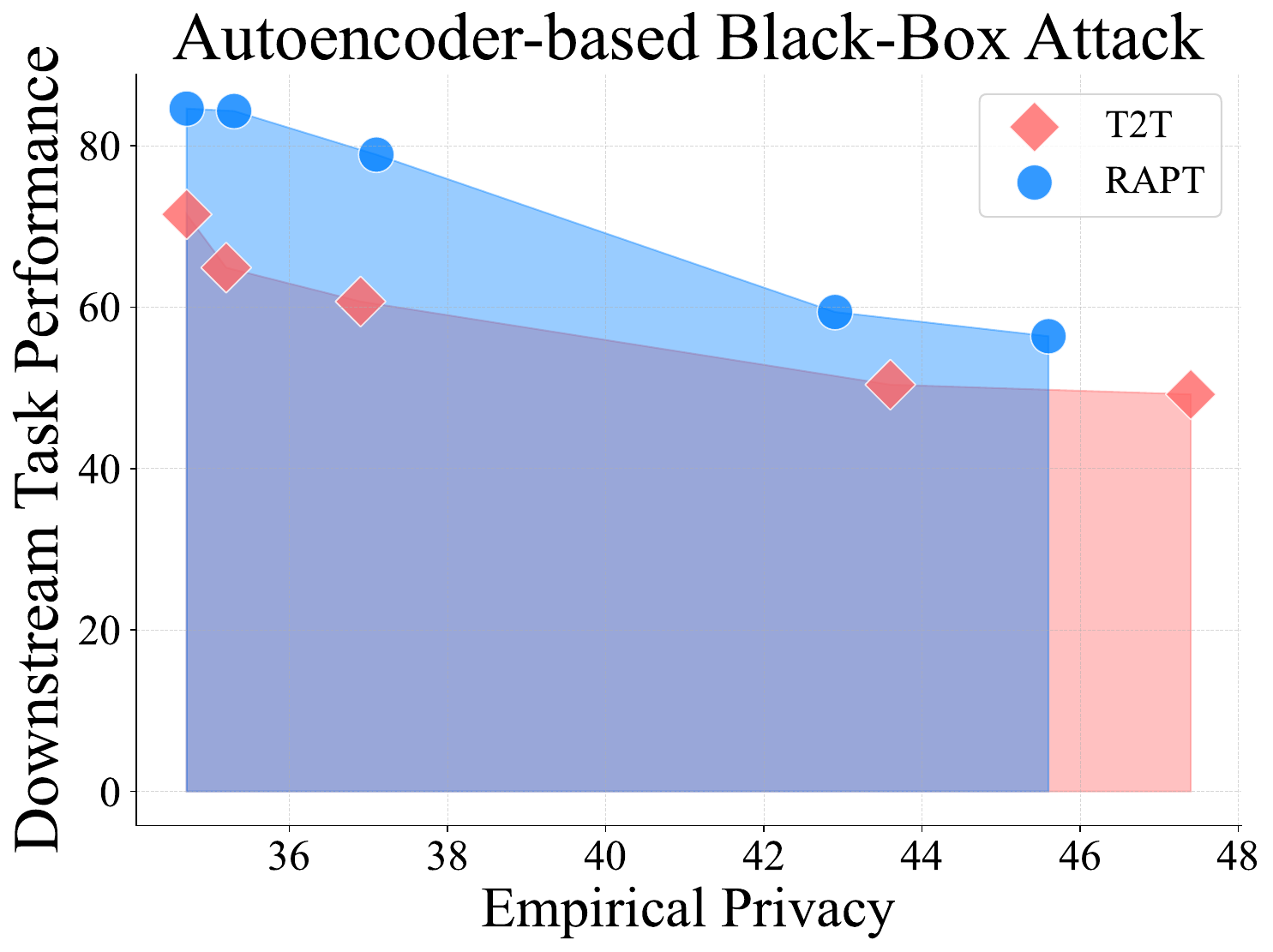}}%
    \subfloat[]{\includegraphics[width=0.25\textwidth]{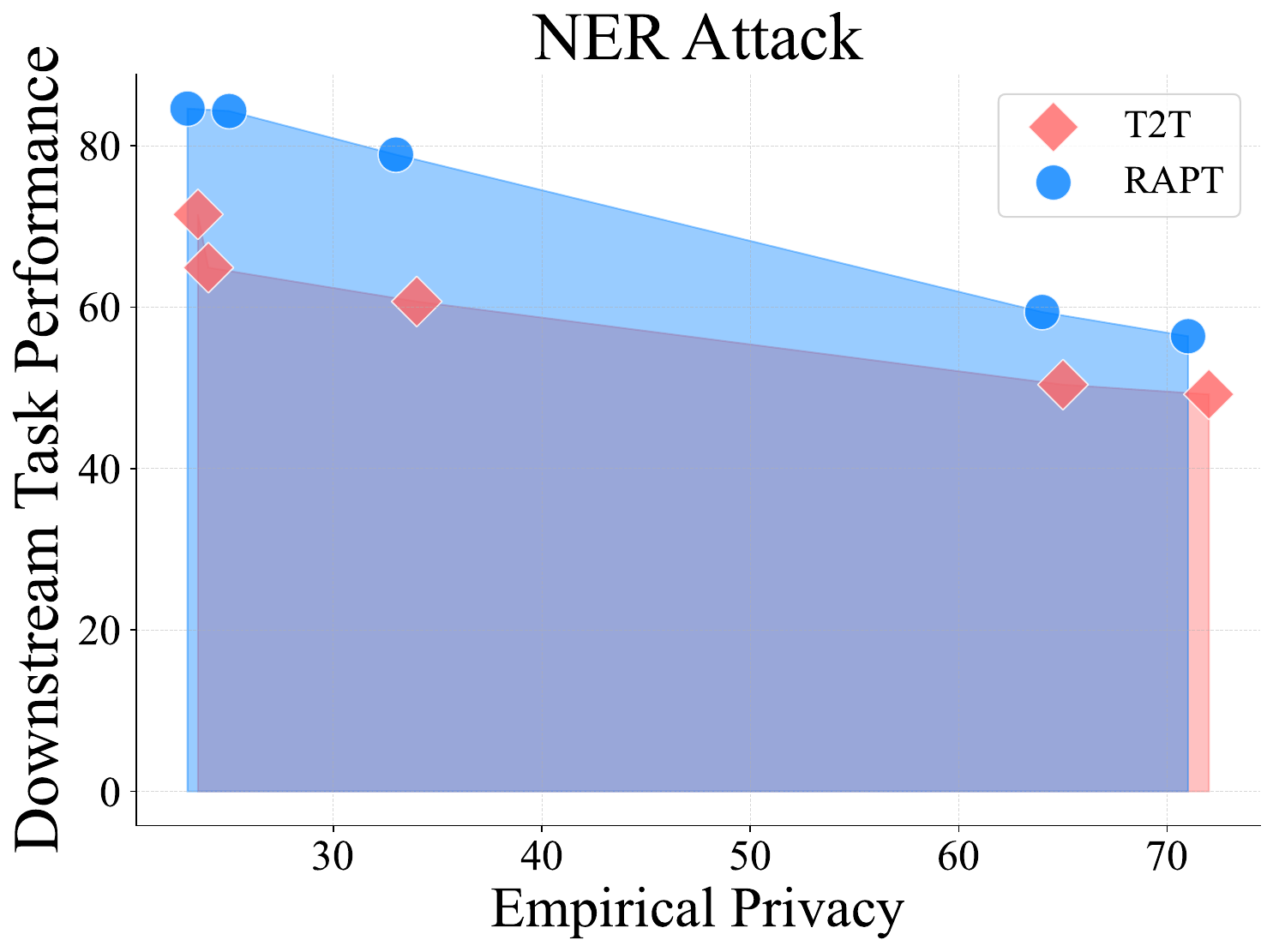}}
    \caption{Pareto frontiers for \textsc{rapt} and T2T privatization against four simulated attacks. The dashed line in each graph represents the results from experiments conducted without the use of privacy protection.}
    \label{fig:Pareto_Frontiers_TR}
\end{figure}

We analyze the Pareto frontiers of our proposed \textsc{rapt} and \textsc{prompt tuning} with T2T privatization on the SST-2 task. We simulate white-box attacks, MLP-based black-box attacks, autoencoder-based black-box attacks, and NER attacks. The backbone LLM is \texttt{BERT}$_{\text{BASE}}$, and the privacy parameter $\eta$ ranges from 25 to 175. The results are illustrated in Figure~\ref{fig:Pareto_Frontiers_TR}. From the figure, we find that for all attacks, \textsc{rapt} consistently outperforms T2T privatization under the same degree of privacy protection. As a result, it is clear that \textsc{rapt} can achieve a better privacy-utility trade-off than using T2T privatization.

\subsection{Overheads on Edge-devices}
\begin{table}[t]
\centering
\caption{Computational overhead introduced by \textsc{rapt} on edge devices for different input lengths.}
\label{tab:edge_overhead}
\resizebox{0.47\textwidth}{!}{
\begin{tabular}{ccccc}
\toprule
\textbf{Length} (words) & \textbf{Time Per Sequence} (s) & \textbf{Peak Memory Usage} (MB)  \\
\midrule
10 & 0.03 & 96.72  \\
100 & 0.05 & 98.02 \\
1,000 & 0.15 & 99.62 \\
10,000 & 0.81 & 100.72 \\
\bottomrule
\end{tabular}}
\end{table}

Local privacy settings require data privatization on the user side and may therefore introduce additional overhead. To analyze the overhead of \textsc{rapt}, we study its execution time and memory consumption when privatizing user inputs on edge devices. We conduct experiments on a Raspberry Pi 4 device, which features four Cortex-A72 cores at 1.5GHz and 8GB of RAM, and use the embedding matrix of \texttt{BERT$_\textrm{BASE}$} for nearest neighbor search. The results are shown in Table~\ref{tab:edge_overhead}. For initialization steps such as loading the embedding matrix and preparing POS mappings, the setup takes approximately 14.3 seconds on the Raspberry Pi 4. For sequences ranging from 10 to 10,000 words, we find that the processing time per sequence ranges from 0.03 to 0.81 seconds, which is acceptable for typical use cases. For memory consumption, processing a sequence of length 10 occupies 96.72MB of memory, of which 89.42MB is taken up by the embedding matrix of the \texttt{BERT$_\textrm{BASE}$} model. When the sequence length is increased to 50,000, the memory usage rises to 100.72MB, adding only 4MB compared to a length of 10, indicating that privatization is not memory-intensive. Therefore, we conclude that the overhead of performing privatization locally is acceptable for most real-world applications.

\subsection{Theoretical Analysis}
Empirical results demonstrate that \textsc{rapt} consistently outperforms baseline methods with privatized token reconstruction. In this subsection, we provide a theoretical justification, showing that joint prediction and denoising are essential for learning the Bayes-optimal predictor when inputs are corrupted by noise.

Let $(\bm{w}, y)$ be random variables with joint distribution $P(\bm{w}, y)$, and let $\hat{\bm{w}} = \bm{w} + \bm{z}$ denote a noisy observation, where the noise $\bm{z}$ is independent of both $\bm{w}$ and $y$. We assume all relevant conditional expectations are well-defined and finite. The following theorem characterizes the decomposition of the Bayes-optimal predictor.

\begin{theorem}[Decomposition of the Bayes-Optimal Predictor]
Let $g^\star(\hat{\bm{w}}) \doteq \mathbb{E}[y \mid \hat{\bm{w}}]$
be the Bayes-optimal predictor under mean squared error. Then
\[
g^{\star}(\hat{\bm{w}}) = f^{\star}\!\big(h^{\star}(\hat{\bm{w}})\big),
\]
where $h^\star(\hat{\bm{w}}) \doteq P(\bm{w} \mid \hat{\bm{w}})$, and $f^\star(\bm{w}) \doteq \mathbb{E}[y \mid \bm{w}]$.
\end{theorem}

\begin{proof}
By definition of the Bayes-optimal predictor under mean squared error,
\[
g^\star(\hat{\bm{w}}) = \mathbb{E}[y \mid \hat{\bm{w}}].
\]
Applying the law of iterated expectation conditional on $\hat{\bm{w}}$ yields
\[
\mathbb{E}[y \mid \hat{\bm{w}}]
= \mathbb{E}\!\left[\mathbb{E}[y \mid \bm{w}] \,\middle|\, \hat{\bm{w}}\right]
= \int \mathbb{E}[y \mid \bm{w}] \, P(\bm{w} \mid \hat{\bm{w}}) \, d\bm{w}.
\]
By the definitions of $h^\star$ and $f^\star$, the right-hand side equals
$f^\star(h^\star(\hat{\bm{w}}))$. Thus,
\[
g^\star(\hat{\bm{w}}) = f^\star(h^\star(\hat{\bm{w}})).\qedhere
\]
\end{proof}
This result implies that learning an optimal predictor from noisy inputs inherently requires estimating the clean input distribution $P(\bm{w} \mid \hat{\bm{w}})$ (denoising) and the clean-input predictor $\mathbb{E}[y \mid \bm{w}]$ (prediction). \textsc{rapt} addresses both objectives jointly via privatized token reconstruction, leading to improved utility despite noise.

\section{Conclusion}
In this work, we proposed \textsc{rapt}, a privacy-preserving parameter-efficient fine-tuning framework for large language model services. \textsc{rapt} introduces POS-constrained text-to-text privatization on the user side and customizes LLM services with PEFT methods and privatized token reconstruction on the service provider side. Empirical results demonstrate that \textsc{rapt} achieves a better privacy–utility trade-off across diverse LLMs and tasks. A limitation of \textsc{rapt} is that it inevitably adds more complexity and overhead to both the user and service provider sides and cannot fully eliminate the performance drop when imposing stronger privacy protection. We plan to investigate these issues in future work.

\bibliography{custom}
\bibliographystyle{IEEEtranN}
\end{document}